\pdfoutput=1

\documentclass[11pt]{article}

\usepackage{pgfplotstable}
\usepackage{multirow}
\usepackage{enumerate}
\usepackage[shortlabels]{enumitem}

\usepackage[final]{acl}
\usepackage{subcaption}

\usepackage{times}
\usepackage{latexsym}
\usepackage{booktabs}

\usepackage[T1]{fontenc}

\usepackage[utf8]{inputenc}

\usepackage{microtype}

\usepackage{inconsolata}

\usepackage{graphicx}
\usepackage{subcaption}
\usepackage{array}

\usepackage[utf8]{inputenc}
\usepackage{tcolorbox}
\usepackage{graphicx}
\usepackage{pgf}

\tcbset{
  colframe=black,
  colback=white,
  fonttitle=\bfseries,
  sharp corners,
  boxrule=0.5pt,
  width=\columnwidth,
  left=4pt,
  right=4pt,
  top=4pt,
  bottom=4pt
}

\usepackage{svg}        
\usepackage{caption}    
\usepackage{subcaption}    
\usepackage{tikz}
\usetikzlibrary{shapes.callouts}
\usetikzlibrary{positioning}
\usepackage{fancyvrb}      
\usepackage{framed}        
\usepackage{xcolor}        
\usepackage{graphicx}      
\usepackage{caption}       
\usepackage{amsmath}  
\definecolor{lightbrown}{rgb}{0.71, 0.40, 0.11}
\definecolor{darkblue}{rgb}{0.00, 0.20, 0.60}
\definecolor{lightblue}{rgb}{0.20, 0.60, 0.86}

\newcommand{\lightbrowntext}[1]{\textcolor{lightbrown}{#1}}
\newcommand{\darkbluetext}[1]{\textcolor{darkblue}{#1}}
\newcommand{\lightbluetext}[1]{\textcolor{lightblue}{#1}}
\definecolor{darkgreen}{rgb}{0, 0.5, 0}
\definecolor{darkred}{rgb}{0.7, 0, 0.1}
\definecolor{darkblue}{rgb}{0, 0, 0.5}

\newcommand{\labeledentity}[3]{%
  \tikz[baseline=(word.base)]{
    \node[anchor=south, inner sep=2pt, fill=#3, rounded corners=2pt, text=white, font=\scriptsize] (label) {\tiny #1};
    \node[below=1pt of label, draw, thick, rounded corners=2pt, inner sep=2pt, font=\small] (word) {#2};
  }%
}

\title{Entity Framing and Role Portrayal in the News}

\author{Tarek Mahmoud$^1$, 
  \textbf{Zhuohan Xie}$^1$,
  \textbf{Dimitar Dimitrov}$^2$,
  \textbf{Nikolaos Nikolaidis}$^3$, \\
  \textbf{Purificação Silvano}$^4$, 
  \textbf{Roman Yangarber}$^5$,
  \textbf{Shivam Sharma}$^6$,
  \textbf{Elisa Sartori}$^{7}$, \\
  \textbf{Nicolas Stefanovitch}$^{8}$, 
  \textbf{Giovanni Da San Martino}$^{7}$,
  \textbf{Jakub Piskorski}$^9$,
  \textbf{Preslav Nakov}$^1$ \\
$^1$MBZUAI, UAE \quad
$^2$Sofia University "St. Kliment Ohridski", Bulgaria \\
$^3$Athens University of Economics and Business, Greece \quad
$^4$University of Porto, Portugal \\
$^5$University of Helsinki, Finland \quad
$^6$Indian Institute of Technology Delhi, India \\
$^7$University of Padova, Italy \quad
$^8$European Commission Joint Research Centre, Italy \\
$^9$Institute of Computer Science, Polish Academy of Sciences, Poland \\
\texttt{\{tarek.mahmoud, preslav.nakov\}@mbzuai.ac.ae}
}

\pgfplotsset{compat=1.18} 
\begin{document}
\maketitle
\begin{abstract}
We introduce a novel multilingual hierarchical corpus annotated for entity framing and role portrayal in news articles.\footnote{Our dataset is available at\\ \url{https://mbzuai-nlp.github.io/entity-framing/}} The dataset uses a unique taxonomy inspired by storytelling elements, comprising 22 fine-grained roles, or archetypes, nested within three main categories: \emph{protagonist}, \emph{antagonist}, and \emph{innocent}. Each archetype is carefully defined, capturing nuanced portrayals of entities such as guardian, martyr, and underdog for protagonists; tyrant, deceiver, and bigot for antagonists; and victim, scapegoat, and exploited for innocents. The dataset includes 1,378 recent news articles in five languages (Bulgarian, English, Hindi, European Portuguese, and Russian) focusing on two critical domains of global significance: the Ukraine--Russia War and Climate Change. Over 5,800 entity mentions have been annotated with role labels. This dataset serves as a valuable resource for research into role portrayal and has broader implications for news analysis. We describe the characteristics of the dataset and the annotation process, and we report evaluation results on fine-tuned state-of-the-art multilingual transformers and hierarchical zero-shot learning using LLMs at the level of a document, a paragraph, and a sentence. 
\end{abstract}

\begin{figure*}[!htbp]
\centering
\footnotesize
\fbox{%
     \parbox{1\textwidth}{
     {\centering \textbf{Putin says what Russia needs to do to win special operation in Ukraine } \\[1em]}

    Russia will win the special operation in Ukraine if the society shows consolidation and composure to the enemy, President Vladimir Putin said during a visit to the Ulan-Ude Aviation Plant on March 14, Rossiya 24 TV channel said.
    
    Russia is not improving its geopolitical position in Ukraine. Instead, \labeledentity{Underdog}{Russia}{darkblue} is fighting "for the survival of Russian statehood, for the future development of the country and our children."
    
    "In order to bring peace and stability closer, we, of course, need to show the consolidation and composure of our society. When the enemy sees that our society is strong, internally braced up, consolidated, then, without any doubt we will come to reach what we are striving for — both success and victory," Putin said.
    
    According to him, many of the current problems began after the collapse of the Soviet Union, when they tried to put pressure on \labeledentity{Victim}{Russia}{darkgreen} to "destabilise the internal political situation.” "Hordes of international terrorists" new sent to the purpose to accomplish this goal, Putin said.
    
    Afterwards, the West decided to start rehabilitating Nazism in Russia's neighbouring states, including in Ukraine.
    
    Nevertheless, Putin continued, Russia had long tried to build partnerships with both Western countries and Ukraine. However, after 2014, when the West contributed to the coup in Ukraine, the state of affairs changed dramatically. It was then when they started exterminating those who advocated the development of normal relations with Russia, he said.
    
    According to Putin, \labeledentity{Guardian}{Russia}{darkblue} was forced to launch the special operation to protect the population. \labeledentity{Saboteur}{Western countries}{darkred} were hoping to break Russia quickly, but they were wrong, he said adding that \labeledentity{Virtuous}{Russia}{darkblue} managed to raise its economic sovereignty since 2022.
    
    }%
}
\caption{Annotated example color-coded according to the main roles in the taxonomy: \textcolor{darkred}{red} for \emph{antagonist}, \textcolor{darkblue}{blue} for \emph{protagonist}, and \textcolor{darkgreen}{green} for \emph{innocent}.}
\label{fig:annotated_example_text2}
\end{figure*}

\section{Introduction\label{sec:intro}}
The rapid proliferation of massive amounts of content on social media has dramatically transformed the information landscape, providing immediate access to news, and allowing anyone to propagate their narratives across the globe. While this connectivity functions as a convenient avenue for information dissemination in a very democratic way, it also heightens the risk of online users of being exposed to biased reporting, propaganda, and narrative manipulation. 

These risks are particularly pronounced during periods of conflict and political upheavals, where the framing of entities—individuals, organizations, or groups—can profoundly influence public perception and decision-making. Understanding how entities are portrayed in the news is essential for fostering media literacy, identifying bias, and ensuring transparent news consumption.

Social science highlights the role of emotion in framing—selecting elements that evoke affective responses to shape perceptions \cite{https://doi.org/10.1111/j.1467-9221.2004.00354.x}. Emotional framing often leverages language that elicits specific feelings, such as fear, anger, or compassion, to influence how entities are understood \cite{iyengar_is_1991, nabi_exploring_2003,https://doi.org/10.1111/j.1460-2466.2000.tb02843.x,brader_campaigning_2006}. For instance, referring to a group as ``freedom fighters'' versus ``terrorists'' not only frames their role but also activates distinct emotional reactions. Research has shown that emotional appeals are powerful tools for reinforcing or challenging public attitudes \cite{westen_political_2008,lerner_emotion_2015}. Such framing can manifest through specific linguistic cues and includes the portrayal of entities also defined by \citet{schneider_social_1993} as the \emph{Social Construction of Target Populations}.

Natural language processing research has increasingly been applied to analyze the emotional dimensions of framing \cite{troiano-etal-2023-relationship}, including the identification of sentiment \cite{zhang-etal-2024-sentiment,app13074550} and emotion-laden narratives \cite{mousavi-etal-2022-emotion} in text. Understanding these emotional components provides deeper insight into how media narratives construct and perpetuate particular representations of entities, which ultimately shapes public perception and societal discourse.

Given the large scale and complexity of the modern news ecosystem, effective analysis of entity framing requires automated tools, which depend on high-quality annotated data. 

In this context, we introduce a new multilingual dataset designed to develop tools for the study of entity framing and role portrayal in news articles. Our dataset uses a unique, hierarchical taxonomy inspired by elements of storytelling containing a set of 22 carefully defined archetypal roles nested under three main frames: \emph{protagonist}, \emph{antagonist}, and \emph{innocent}.

The corpus spans 1,378 recent news articles in five languages (Bulgarian, English, Hindi, European Portuguese, and Russian) and focuses on two globally significant domains: the Ukraine-Russia war and climate change. We annotated over 5,800 entity mentions with detailed role labels.

Our contributions can be summarized as follows:

\begin{itemize}
    \item We release a novel multilingual dataset annotated for entity framing and role portrayal, complete with detailed annotation guidelines.
    \item We introduce a comprehensive hierarchical taxonomy for entity roles validated on a large set of documents, supporting analysis at both the coarse and the fine-grained levels.
    \item We provide comprehensive dataset statistics, and analyze co-occurrence of roles as well as role transitions.
    \item We set benchmarks using state-of-the-art multilingual transformer models, and hierarchical zero-shot learning with LLMs.
\end{itemize}

\section{Related Work\label{sec:related_work}}

\subsection{A Hierarchical Taxonomy of Entity Roles}

The identification of narrative frames such as heroes, villains, and victims has been studied across various modalities and domains. \citet{sharma-etal-2023-characterizing} introduced a dataset for identifying heroes, villains, and victims in memes, focusing on visual content. Similarly, \cite{stammbach-etal-2022-heroes} demonstrated that LLMs can detect these frames from narrative texts using zero-shot question-answering prompts across diverse domains such as newspaper articles, movie plot summaries, and political speeches. \citet{frermann-etal-2023-conflicts} integrated elements like conflict and resolution with the framing of key entities as heroes, victims, or villains. 

These studies primarily focus on the coarse-grained categorization of entities into the hero-villain-victim frames. \citet{doi:10.1177/0190272518781050} analyze this triad and show how these frames are imbued with moral attributes where heroes are shown as benevolent and strong, villains as malevolent and strong, and victims as benevolent and weak. 
In contrast, our framework redefines these frames based on narrative function, rather than moral judgment. By introducing the frames of protagonist, antagonist, and innocent, we focus on the functional roles entities occupy within a narrative, independent of their moral aspect. 

This approach allows for a more nuanced and flexible analysis of entity roles. We further present a hierarchical taxonomy of 22 roles nested within these narrative functional frames. This rich set of roles captures the diverse ways entities are portrayed in textual narratives. Our taxonomy was validated through human annotations across diverse news domains.

\subsection{The Entity Framing Task Formulation}

\paragraph{Aspect-Based Sentiment Analysis (ABSA)} Prior work on ABSA and entity-level sentiment analysis, which is a special case of ABSA and the most relevant one to our research, explores both the assignment of sentiment to particular attributes of entities and to entities as a whole \cite{saeidi-etal-2016-sentihood, jiang-etal-2019-challenge, orbach-etal-2021-yaso, DBLP:journals/corr/abs-2203-01054, chebolu-etal-2024-oats, kuila-sarkar-2024-deciphering, bastan-etal-2020-authors, ronningstad-etal-2022-entity}. For instance, \citet{tang-etal-2023-finentity} presented a dataset for entity-level sentiment classification in financial texts. In contrast, our task formulation differs by not assigning polarities or defining aspects; instead, we classify entities into a rich set of roles inspired by storytelling elements.

\paragraph{Article-level framing} \citet{card-etal-2016-analyzing} addressed article-level framing by developing a model that utilizes personas to infer framing dimensions defined in the Media Frames Corpus \cite{card-etal-2015-media}. Their topic modeling approach identified 50 personas, but only a few, such as ``refugee'' and ``immigrant,'' were found to be informative. Further research on news framing has predominantly focused on article-level analysis. Studies like \citet{liu-etal-2019-detecting, piskorski-etal-2023-multilingual, DBLP:conf/acl/0001KF24, Pastorino2024DecodingNN} have explored various dimensions of framing in news articles. However, our approach centers on the framing of individual entities within the text, not on article-level framing.

\section{The Entity Framing Task}
Entity framing focuses on analyzing how a text portrays a specific entity through word choice and narrative structure. More concretely, given a news article and a list of entity mentions (i.e., entity mentions, along with their span offsets), we assign to each of them one or more roles based on the taxonomy shown in Figure \ref{fig:taxonomy_roles}. We developed this taxonomy specifically for this task and the roles were inspired by storytelling elements. 

The taxonomy includes 22 archetypes, or fine-grained roles nested under three main frames: \emph{protagonist}, \emph{antagonist}, and \emph{innocent}. The role an entity plays in a given article may differ from one context in that article to another depending on the portrayal. See Figure~\ref{fig:annotated_example_text2} for a complete, annotated example from the corpus.

Entity framing can be formalized mathematically as follows. Let $R$ be a tree structure representing the taxonomy of roles. Let $S$ be a string of length $|S|$ characters with the content of the full article. The goal of entity framing is to learn a function
\begin{equation}
f: (S, [i,j]) \rightarrow \{r_1, r_2, \ldots, r_k\} \subseteq R
\end{equation}

\noindent where $0 \leq i < j \leq |S|$ and $\{r_1, r_2, \ldots, r_k\}$ is the set of roles assigned to the span $[i,j]$.


\begin{figure}[!ht]
    \centering
\begin{tcolorbox}
\scriptsize

\textbf{PROTAGONIST}
\medskip

\textbf{Guardian:} Heroes or guardians who protect values or communities, ensuring safety and upholding justice.

\noindent\textbf{Martyr:} Individuals who sacrifice their well-being, or even their lives, for a greater good or cause.

\noindent\textbf{Peacemaker:} Individuals who advocate for harmony, resolving conflicts and bringing about peace.

\noindent\textbf{Rebel:} Revolutionaries who challenge the status quo and fight for significant change or liberation.

\noindent\textbf{Underdog:} Entities who, despite a disadvantaged position, strive against greater forces and obstacles.

\noindent\textbf{Virtuous:} Individuals portrayed as righteous, fair, and upholding high moral standards.

\medskip

\noindent\textbf{ANTAGONIST}

\medskip

\noindent\textbf{Instigator:} Those who initiate conflict and provoke violence or unrest.

\noindent\textbf{Conspirator:} Individuals involved in plots and covert activities to undermine or deceive others.

\noindent\textbf{Tyrant:} Leaders who abuse their power, ruling unjustly and oppressing others.

\noindent\textbf{Foreign Adversary:} Entities from other nations creating geopolitical tension and acting against national interests.

\noindent\textbf{Traitor:} Individuals who betray a cause or country, seen as disloyal and treacherous.

\noindent\textbf{Spy:} Individuals engaged in espionage, gathering and transmitting sensitive information.

\noindent\textbf{Saboteur:} Those who deliberately damage or obstruct systems to cause disruption.

\noindent\textbf{Corrupt:} Individuals or entities engaging in unethical or illegal activities for personal gain.

\noindent\textbf{Incompetent:} Entities causing harm through ignorance, lack of skill, or poor judgment.

\noindent\textbf{Terrorist:} Individuals who engage in violence and terror to further ideological ends.

\noindent\textbf{Deceiver:} Manipulators who twist the truth, spread misinformation, and undermine trust.

\noindent\textbf{Bigot:} Individuals accused of hostility or discrimination against specific groups.

\medskip

\noindent\textbf{INNOCENT}

\medskip

\noindent\textbf{Forgotten:} Marginalized groups who are overlooked and ignored by society.

\noindent\textbf{Exploited:} Individuals or groups used for others’ gain, often without consent.

\noindent\textbf{Victim:} People suffering harm due to circumstances beyond their control.

\noindent\textbf{Scapegoat:} Entities unjustly blamed for problems or failures to divert attention.
\end{tcolorbox}

    \caption{The hierarchical taxonomy of roles for entity framing. A more comprehensive description accompanied with examples is provided in Appendix \ref{sec:detailed_taxonomy}.}
    \label{fig:taxonomy_roles}
\end{figure}

\section{Corpus Description}

\subsection{Domains}
\label{sec:corpus_domains}
The articles used in the task cover the following domains: (1) \emph{Ukraine-Russia War}, which includes articles about the war that started in February 2022 when Russia launched a full-scale invasion of Ukraine and began occupying parts of the country, and (2) \emph{Climate Change}, which encompasses both climate change denial (characterized by rejecting, refusing to acknowledge, disputing, or fighting the scientific consensus on climate change), and climate change activism.

\subsection{Article Selection}

For each language, articles were primarily selected from links we obtained from a large-scale in-house news aggregation tool. We performed the first candidate article selection based on multilingual keyword-based filters and we perform several steps, which we follow by to enrich the selection to match the criteria discussed below. To select the articles, we followed these steps:

\paragraph{Initial Collection:}  
Articles were scraped and filtered based on criteria such as word count (e.g.,~only articles exceeding 250 words were selected). For duplicate articles, the version with the higher number of words was preferred.

\paragraph{Filtering:}  
Each article was manually reviewed to determine its relevance to the annotation task. The articles were categorized into four groups: Perfect Fit, Average Fit, Uncertain (requiring further validation by language coordinators), or Unfit (excluded from annotation). Only articles classified as \emph{Perfect Fit} and \emph{Average Fit} were considered for annotation. 
Afterwards, we used a zero-shot classifier with selected key phrases, and a persuasiveness score using the persuasion technique classifier as described in \citet{nikolaidis-etal-2024-exploring}. These scores were used to further enrich the selection with relevant articles.

Additional sources were also incorporated to ensure diversity of perspectives for two languages: for Hindi, we selected articles from mainstream and bias-specific outlets (e.g., NDTV, The Hindu, OpIndia), and for Portuguese, from newspapers and political websites (e.g., \emph{O Diabo}, Esquerda, Folha Nacional) that had more controversial opinion texts about the relevant topics.

\subsection{Annotation Process}

Given that our corpus contains articles in five languages, we had separate annotation teams, one for each language, each led by a language coordinator. Each language team included 3 to 5 annotators with prior experience in linguistics, social science, international relations, or prior annotation work. The annotators studied our detailed guidelines, attended live demonstrations, and completed real-time annotation exercises. During weekly meetings, teams clarified any uncertainties, resolved conflicts, improved consistency, and revised the annotation guidelines.

Each article was annotated by two annotators. Designated curators, often language coordinators or experienced annotators, reviewed and consolidated all annotations. They resolved the discrepancies through discussions with the respective teams. Over time, as the annotation quality improved, the curators reduced the frequency of checks but continued to perform random quality checks to ensure that annotations were of high quality. We used the Inception tool~\cite{tubiblio106270} for annotating and curating the corpus. See Appendix~\ref{sec:annotation_tool} for more details.


The annotation guidelines (Appendix \ref{sec:annotation_guidelines}) were refined during initial weekly meetings between language coordinators and annotators. From these guidelines, we emphasize key aspects of entity selection for annotation. We annotated traditional named entities, extending this to also include eponym-derived entities (e.g., \emph{Putin supporters}) and toponym-derived entities (e.g., \emph{Western countries}, as illustrated in Figure~\ref{fig:annotated_example_text2}). Additionally, we focused on annotating entities that are central to the narrative conveyed by the article. For example, in Figure~\ref{fig:annotated_example_text2}, entities such as \emph{Ulan-Ude Aviation Plant} and \emph{Rossiya 24 TV} were not annotated because they were not considered central to the narrative. For a detailed explanation of how centrality was defined, refer to the guidelines in Appendix~\ref{sec:annotation_guidelines}.

\begin{table*}[ht]
\small
\centering
\resizebox{0.9\textwidth}{!}{%
\begin{tabular}{c|rrrrrrrrr|rrrr}
\toprule
\multicolumn{1}{c|}{Lang.} & 
\multicolumn{1}{c}{\#DOC} & 
\multicolumn{1}{c}{\#PAR} & 
\multicolumn{1}{c}{\#SEN} & 
\multicolumn{1}{c}{\#WORD} & 
\multicolumn{1}{c}{\#CHAR} & 
\multicolumn{1}{c}{AVG\textsubscript{p}} & 
\multicolumn{1}{c}{AVG\textsubscript{s}} & 
\multicolumn{1}{c}{AVG\textsubscript{w}} & 
\multicolumn{1}{c|}{AVG\textsubscript{c}} & 
\multicolumn{1}{c}{\#ENT} & 
\multicolumn{1}{c}{\#ANN} & 
\multicolumn{1}{c}{AVG\textsubscript{e}} & 
\multicolumn{1}{c}{AVG\textsubscript{a}}\\
\midrule
BG    & 274    & 2.7K  & 5.0K  & 104K  & 584K   & 9.8  & 18.5 & 380.6 & 2129.8 & 656 (179)     & 742   & 2.4 & 2.7 \\
EN    & 229    & 3.7K  & 5.3K  & 131K  & 705K   & 16.2 & 23.1 & 571.3 & 3080.7 & 775 (413)     & 843   & 3.4 & 3.7 \\
HI    & 377    & 3.8K  & 8.7K  & 200K  & 947K   & 10.2 & 23.0 & 530.1 & 2513.0 & 2,609 (724)   & 3,030 & 6.9 & 8.0 \\
PT    & 337    & 3.5K  & 5.4K  & 150K  & 822K   & 10.5 & 15.0 & 445.0 & 2439.2 & 1,365 (440)   & 1,438 & 4.1 & 4.3 \\
RU    & 161    & 0.7K  & 2.0K  & 42K   & 257K   & 4.5  & 12.7 & 261.1 & 1599.1 & 451 (265)     & 477   & 2.8 & 3.0 \\\midrule
Total & 1,378  & 14.5K & 26.4K & 627K  & 3,316K & 10.5 & 19.2 & 455.0 & 2406.3 & 5,856 (1,962) & 6,530 & 4.2 & 4.7 \\
\bottomrule
\end{tabular}
}
\caption{Corpus statistics showing total number of documents (\#DOC), paragraphs (\#PAR), sentences (\#SEN), words (\#WORD), and characters (\#CHAR) by language. The averages (AVG\textsubscript{p}), (AVG\textsubscript{s}), (AVG\textsubscript{w}), and (AVG\textsubscript{c}) refer to the average number of paragraphs, sentences, words, and characters per document, respectively. The table also shows the total number of annotated entity mentions (\#ENT) accompanied with unique counts in parentheses, the total number of annotations (\#ANN), the average number of entity mentions per document (AVG\textsubscript{e}), and the average number of annotations per document (AVG\textsubscript{a}).}
\label{tab:corpus_stats}
\end{table*}


\begin{table}[!h]
\small
    \centering
\begin{tabular}{ccccccc}
\toprule
Lang. & EN & RU & BG & PT & HI & All\\
\hline
$\alpha$ & .460 & .436 & .733 & .467 & .461 & .558\\
\bottomrule
\end{tabular}
    \caption{Inter-annotator agreement computed using Krippendorff's $\alpha$.}
    \label{tab:iaa}
\end{table}

\begin{table}[!h]
    \small
    \centering
    \resizebox{\columnwidth}{!}{%
    \begin{tabular}{lllllll}
        \toprule
        \textbf{Freq.} & \textbf{1} & \textbf{2-5} & \textbf{5-10} & \textbf{10-20} & \textbf{20-50} & \textbf{50-500} \\
        \midrule
        count (\%) & 1513 (74) & 374 (18) & 76 (4) & 46 (2) & 22 (1) & 14 (1) \\
\bottomrule
\end{tabular}
}
\caption{Proportion of entities of a given frequency.}
    \label{tab:proportion}
\end{table}


\subsection{Inter-Annotator Agreement}

To assess the inter-annotator agreement (IAA), we used Krippendorff's alpha. We compared the annotations at the span level: they were approximately matched if they shared at least 50\% of their length, to account for minor differences.
The IAA values are shown in Table~\ref{tab:iaa}. The results indicate a moderate agreement (above 0.45), which we consider acceptable due to the span-based nature of the task. 

We can see that the IAA is similar across the languages, except for Bulgarian, for which it is notably higher (0.73), which can be explained by the low count of entities in Table~\ref{tab:corpus_stats}.  
\

\subsection{Corpus Analysis}

\subsubsection{Statistics}

\begin{figure*}[htbp]
    \centering
    \includegraphics[width=0.95\textwidth]{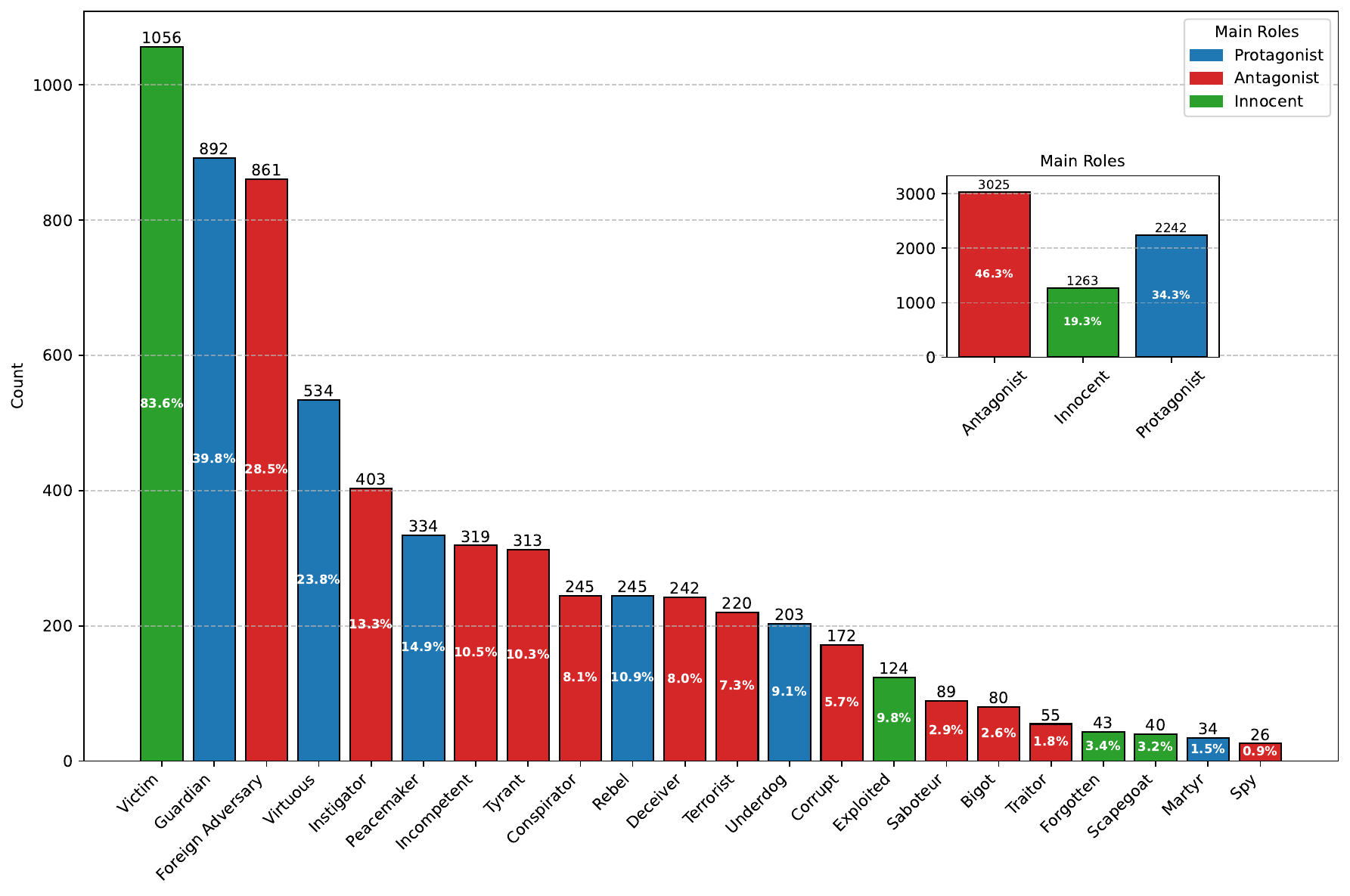}
    \caption{Distribution of fine-grained roles color-coded according to the main role. For the fine-grained roles, the percentages inside the bars indicate the proportion of each fine-grained role relative to its corresponding main role category. The counts on top of the bars show the total occurrences of each fine-grained role. In the mini histogram, the percentages inside the bars reflect the distribution of the main roles, with the counts displayed above the bars. }
    \label{fig:fine_roles_histogram}
\end{figure*}

Table \ref{tab:corpus_stats} provides overall statistics about the corpus, including a breakdown per language, the number of annotated entity mentions, the number of unique entities, as well as the number of annotations. Figure~\ref{fig:fine_roles_histogram} displays the distribution of the main and fine-grained roles in the corpus. We can see that there is a significant imbalance between the fine-grained roles, while the distribution of the main roles is relatively balanced. Notably, within the \emph{innocent} category, the majority of the instances, $83.6\%$, are labeled as \emph{victim}, with fewer occurrences of \emph{exploited}, \emph{forgotten}, and \emph{scapegoat} roles. More details about the proportions of main roles and fine-grained roles across different languages can be found in Figure~\ref{fig:proportions_of_roles} in Appendix~\ref{sec:appendix_stats}. 

\begin{figure}
    \centering
    \includegraphics[width=\linewidth]{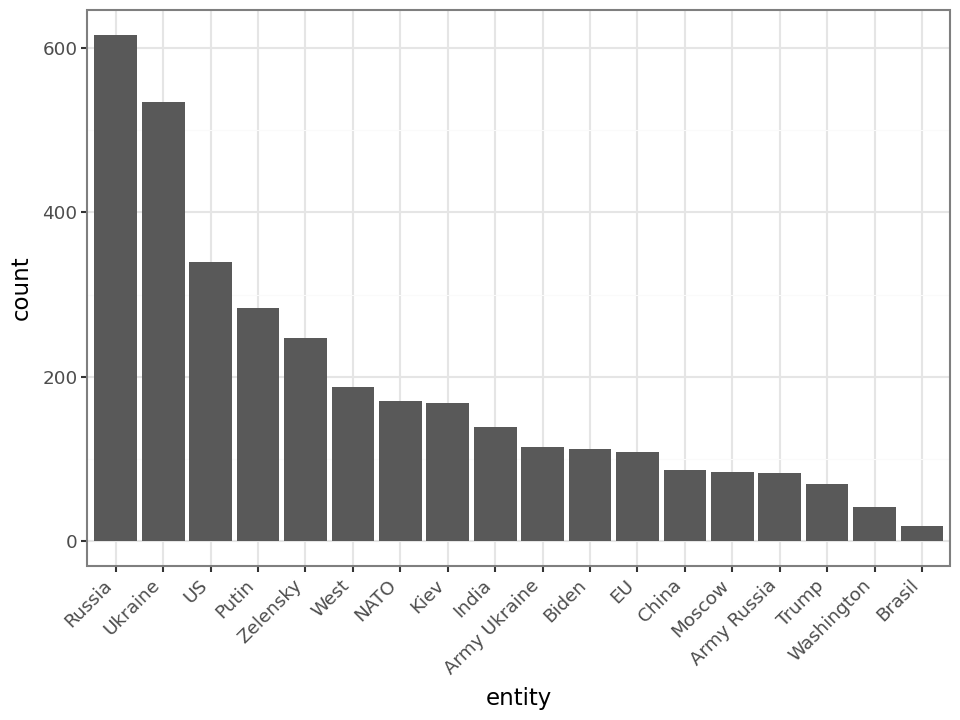}
    \caption{Top entities' counts after multilingual linking was manually performed to link surface string to unique identifiers.}
    \label{fig:hist_ent}
\end{figure}

Table~\ref{tab:proportion} shows the number and proportion of entities within a given frequency range based on exact string matching. We can see that 74\% of the entities were annotated only once, while 14 entities were annotated more than 50 times.
These numbers only reflect surface string, not accounting for grammatical and name variants, or languages. 

In Figure~\ref{fig:hist_ent}, we match the most frequent entities accounting for these differences and present detailed statistics at the level of entities. The entities selected are all the ones for which at least one surface string has a count of at least 10 in any language.

\subsubsection{Co-occurrence of Roles}

In our definition of entity framing, entity mentions can be assigned one or more fine-grained roles. Our corpus contains on average 1.12 annotations per entity mention. Out of the 1,378 articles, 353 articles contain 638 instances where at least one entity mention has multiple annotations. For this set of articles, the average and the maximum are 2.1 and 3 annotations per entity mention, respectively. We further observe that roles such as \emph{peacemaker} frequently accompany \emph{guardian}. Similarly, entities portrayed as \emph{scapegoats} are often framed as being \emph{exploited}. More details about the co-occurrence matrix for entity mentions with multiple annotations are shown in Figures \ref{fig:fine_roles_co_occurence_normalized} and \ref{fig:frequent_roles_combined} in Appendix~\ref{sec:appendix_stats}.

\subsubsection{Analysis of Role Transitions}

We analyze role transitions of an entity within the same article, and observed that out of 1,378 articles, only 99 contained occurrences with main role changes for the same entity, and 261 contained fine-grained role changes.





While the most common main role transitions are between protagonist and antagonist roles, fine role transitions are more nuanced. Examples include sequences like \textbf{victim} to \textbf{underdog} to \textbf{virtuous} and transitions from \textbf{rebel, and virtuous} to \textbf{exploited and victim}. We study the transition sequences for both main and fine-grained roles, and present the ten most frequent transitions sequences in Figure~\ref{fig:transition_sequences} in Appendix~\ref{sec:appendix_stats}.

Our annotation guidelines account for such instances. While such cases are relatively infrequent, they highlight the complexity of entity framing and the importance of capturing role transitions within our taxonomy. This analysis also reinforces the value of our dataset for studying such nuanced challenges.




\section{Experiments}
We experimented with classifying entities into main roles and fine-grained roles. As we performed the entity framing annotations at the span level, we framed the problem as a multi-class multi-label classification task. Given an article, an entity mention, and the span offsets, the goal was to predict the framing of that entity mention. We provide two sets of baselines and experiments to benchmark state-of-the-art models, as well as to assess the complexity of the entity framing task. The first set evaluates fine-tuning multilingual transformer models in various settings, while the second set explores hierarchical zero-shot learning using LLMs. To prevent data leakage, we created train/dev/test splits based on entire articles rather than individual entity-mention annotations. The details of these splits are provided in Table~\ref{tab:train_dev_test_split}.

\begin{table}[]
\centering
\resizebox{\columnwidth}{!}{%

\begin{tabular}{lccccc|c}
\toprule
 & BG & EN & HI & PT & RU & All \\
\midrule
Train & 165 (389) & 133 (440) & 203 (1347) & 206 (833) & 89 (252) & 796 (3261) \\
Dev & 94 (237) & 69 (245) & 139 (983) & 100 (417) & 44 (114) & 446 (1996) \\
Test & 15 (30) & 27 (90) & 35 (279) & 31 (115) & 28 (85) & 136 (599) \\ \midrule
Total & 274 (656) & 229 (775) & 377 (2609) & 337 (1365) & 161 (451) & 1378 (5856) \\
\bottomrule
\end{tabular}
}
\caption{Distribution of articles and entity mentions by language and split. The number of entity mentions is shown in parentheses}
\label{tab:train_dev_test_split}
\end{table}

\subsection{Fine-Tuning Pre-trained Multilingual Transformers}
For the first set of experiments, we designed a series of experiments to address the following important aspects:
\begin{itemize}
    \item \textbf{Granularity of context}--predicting role labels for entity mentions using the full document or narrowing the model's focus to only look at the pertinent paragraph or sentence containing the entity of interest.
    \item \textbf{Multilingual} comparison of the performance in the monolingual setting versus the multilingual setting trained on data in all five languages: Bulgarian, English, Hindi, European Portuguese, and Russian.    
\end{itemize}

For both granularity-level classification and multilingual performance, we made predictions at two levels: main role (3 labels) and fine-grained role (22 labels). We fine-tuned the multilingual pre-trained transformer XLM-R \cite{conneau2020unsupervisedcrosslingualrepresentationlearning} and adapted its final layers for our tasks, applying $softmax$ for multiclass classification and $sigmoid$ for multi-label classification. To handle spans within potentially long documents, we addressed the 512-token limitation of XLM-R by narrowing the context to the paragraph or the sentence where the entity appeared. We constructed the input text using the following format:

\texttt{input = entity mention + [SEP] + title + [SEP] + context}.

In this setup, \texttt{[SEP]} is the separator token, and the context can vary based on the granularity level, ranging from the full text to just the paragraph or sentence containing the entity mention. We placed the entity mention first, followed by the title and context, to maintain consistent positional encodings for the entity mention across different inputs. We used Stanza \cite{qi2020stanza} for sentence splitting for all languages.


To further support span-level multi-label classification, we modified the output layer to include a $sigmoid$ activation and optimized the model using \emph{Binary Cross-Entropy loss}. This setup allowed the model to predict multiple overlapping roles for a given entity span. See Appendix~\ref{sec:appendix_experiments} for more details.

\subsection{Hierarchical Zero-Shot Learning with LLMs}

We experimented with two prompting approaches: \emph{single-step} and \emph{hierarchical multi-step}. The former aimed to predict both the main role and the fine-grained role simultaneously within a single prompt. It assumed that both tasks can be handled together, relying on the model's ability to process them in one go. On the other hand, the multi-step approach separated the prediction into two distinct stages. First, the main role was predicted, and based on that output, the fine-grained role was predicted in a second step, using the information from the initial prediction to refine the second task. This stepwise process involved an intermediate prediction, which allowed the model to focus on each task individually. See Appendix~\ref{sec:appendix_experiments}, and~\ref{sec:appendix_zeroshot} for more details on experimental settings and prompt structure.


\begin{table*}[ht]
\centering
\resizebox{0.65\textwidth}{!}{%
\begin{tabular}{cccccccc}
\toprule
\multirow{2}{*}{\textbf{Train}} &
  \multirow{2}{*}{\textbf{Context}} &
  \multicolumn{2}{c}{\textbf{Main Role}} &
  \multicolumn{4}{c}{\textbf{Fine Grained Role}} \\ \cline{3-8} 
 &
   &
  \textbf{Accuracy} &
  \textbf{Balanced Accuracy} &
  \textbf{P} &
  \textbf{R} &
  \textbf{Micro F1} &
  \textbf{Macro F1} \\ \hline
\multirow{3}{*}{\textbf{M}} & DOC & .6010          & .5904          & --             & --             & --             & --             \\
                   & PAR & .7379          & .7385          & --             & --             & --             & --             \\
                   & SEN & .7179          & .7123          & --             & --             & --             & --             \\ \midrule
\multirow{3}{*}{\textbf{F}} & DOC & .7229          & .7235          & .3495          & .4446          & .3913          & .2306          \\
                   & PAR & \textbf{.7529} & \textbf{.7553} & .3649          & .4985          & .4213          & .2392          \\
                   & SEN & .7496          & .7503          & \textbf{.4195} & \textbf{.4492} & \textbf{.4339} & \textbf{.2529} \\ \bottomrule
\end{tabular}
}
\caption{Performance of entity framing across different granularity settings using XLM-R trained on the full multilingual dataset. Models are trained and evaluated on texts with varying context sizes: full document (DOC), paragraph (PAR), or sentence (SEN) containing the entity mention. The results cover models trained on main roles (M), fine-grained roles (F), and evaluated on either main roles, fine-grained roles, or both.}
\label{tab:xlmr_granularity_results}
\end{table*}






\begin{table}[h]
\centering
\begin{subtable}[h]{\columnwidth}
\centering
\caption{Monolingual setting}
\resizebox{0.75\columnwidth}{!}{
\begin{tabular}{ccccc}
\toprule
Lang. & P & R & Micro F1 & Macro F1 \\
\midrule
EN & $.1032$ & $.1313$ & $.1156$ & $.0435$ \\
BG & $.1056$ & $.5758$ & $.1784$ & $.0505$ \\
HI & $.3424$ & $.4495$ & $.3887$ & $.1740$ \\
PT & $.6124$ & $.6423$ & $.6270$ & $.1505$ \\
RU & $.1077$ & $.5227$ & $.1786$ & $.0437$ \\
\bottomrule
\end{tabular}
}
\end{subtable}


\begin{subtable}[h]{\columnwidth}
\centering
\caption{Multilingual setting}
\resizebox{0.75\columnwidth}{!}{
\begin{tabular}{ccccc}
\toprule
Lang. & P & R & Micro F1 & Macro F1 \\
\midrule
All & $.3649$ & $.4985$ & $.4213$ & $.2392$ \\ \midrule
EN & $.1854$ & $.2828$ & $\mathbf{.2240}$ & $\mathbf{.1327}$ \\
BG & $.3030$ & $.3030$ & $\mathbf{.3030}$ & $\mathbf{.1349}$ \\
HI & $.3234$ & $.4951$ & $\mathbf{.3912}$ & $\mathbf{.2043}$ \\
PT & $.6259$ & $.7480$ & $\mathbf{.6815}$ & $\mathbf{.2040}$ \\
RU & $.4831$ & $.4886$ & $\mathbf{.4859}$ & $\mathbf{.2364}$ \\
\bottomrule
\end{tabular}
}
\end{subtable}

\caption{Results for multi-label fine-grained role classification with XLM-R trained on monolingual and multilingual data and evaluated at the paragraph level.}
\label{tab:xlmr_language_results}
\end{table}

\subsection{Results}

We report standard evaluation metrics, including \emph{micro-average precision}, \emph{recall}, and $F1$ score, along with the \emph{macro-average} $F1$ score for fine-grained roles to address class imbalance. We further provide \emph{accuracy} and \emph{balanced accuracy} for predictions on the main role granularity.

Table~\ref{tab:xlmr_granularity_results} shows the performance of XLM-R across different context granularities (document, paragraph, and sentence) and training configurations (main roles vs. fine-grained roles). For models trained and evaluated on main roles, paragraph-level contexts perform best, followed closely by sentence-level contexts, while document-level contexts perform the worst.  When models trained on fine-grained roles are evaluated on main roles, paragraph-level contexts again yield the best performance, with document and sentence-level contexts slightly behind. This indicates that training on fine-grained roles provides an advantage. For models trained and evaluated on fine-grained roles, sentence-level contexts perform best, followed by paragraph-level contexts, with document-level contexts showing the weakest performance. These results highlight that context granularity significantly impacts performance, with localized contexts outperforming document-level contexts for both main role and fine-grained role classification tasks.

Table~\ref{tab:xlmr_language_results} offers interesting insights into the performance of XLM-R when fine-tuned on monolingual and multilingual data for multi-label fine-grained role classification at the paragraph level. 

The monolingual setting exhibits varying performance across languages, with the highest scores achieved for Portuguese, while English, Bulgarian, and Russian show notably lower performance; the model's performance on Hindi is moderate. These differences stem from the quantity and quality of training data, linguistic variations, and the complexity of entity mentions across languages. The consistently low Macro F1 scores across all languages indicate difficulty in predicting rare roles. In contrast, the multilingual setting consistently outperforms the monolingual setting, demonstrating its ability to better capture diverse fine-grained roles through cross-lingual transfer.


\begin{table*}[t]
\small
\centering
\centering
\resizebox{0.75\textwidth}{!}{%

\begin{tabular}{ccccccccc}
\toprule
\multirow{2}{*}{\textbf{Method}} & \multirow{2}{*}{\textbf{Lang.}} & \multicolumn{2}{c}{\textbf{Main Role}} & \multicolumn{4}{c}{\textbf{Fine Grained Role}} & \multirow{2}{*}{\textbf{Cost (USD)}} \\
\cmidrule(lr){3-4}
\cmidrule(lr){5-8}
 & &  \textbf{Accuracy} & \textbf{Balanced Accuracy} & \textbf{P} & \textbf{R} & \textbf{Micro F1} & \textbf{Macro F1} &  \\
\midrule
\multirow{5}{*}{\shortstack{\textbf{Single-Step} \\ \textbf{LLM Prompting}}}
& EN & .8346 & .6756 & .2692 & .4632 & .3405 & .2171 & 0.7989 \\
& BG & .8065 & .7380 & .3725 & .5588 & .4471 & .3481 & 0.2751 \\
& HI & .6327 & .6247 & .2753 & .4000 & .3262 & .2196 & 2.4696 \\
& PT & .7812 & .7455 & .5167 & .6643 & .5813 & .2891 & 1.0200 \\
& RU & .7558 & .6719 & .3939 & .5843 & .4706 & .4644 & 0.7587 \\
& All & .7030 & .6957 & \underline{.3211} & \underline{.4726} & \underline{.3824} & \underline{\textbf{.3103}} & 5.3223 \\
\midrule
\multirow{5}{*}{\shortstack{\textbf{Multi-Step} \\ \textbf{LLM Prompting}}} 
& EN & .8031 & .6799 & .2887 & .4118 & .3394 & .2383 & 0.5130 \\
& BG & .8031 & .6799 & .4318 & .5588 & .4872 & .3601 & 0.5130 \\
& HI & .6367 & .6284 & .2676 & .2868 & .2769 & .1771 & 1.4581 \\
& PT & .8125 & .7882 & .3895 & .2643 & .3149 & .2498 & 0.5634 \\
& RU & .7442 & .6680 & .4118 & .4719 & .4398 & .3774 & 0.4769 \\
& All & \underline{.7053} & \underline{.7017} & .3051 & .3294 & .3168 & .2765 & \underline{\textbf{3.1852}} \\
\midrule \midrule
\multirow{5}{*}{\textbf{XLM-R}} 
& EN  & 0.6889 & 0.5276 & .1854 & .2828 & .2240 & .1327 & --\\
& BG  & 0.7333 & 0.5791 & .3030 & .3030 & .3030 & .1349 & --\\
& HI  & 0.7025 & 0.7046 & .3234 & .4951 & .3912 & .2043 & --\\
& PT  & 0.8957 & 0.8840 & .6259 & .7480 & .6815 & .2040 & --\\
& RU  & 0.8000 & 0.7604 & .4831 & .4886 & .4859 & .2364 & --\\
& All & \textbf{0.7529} & \textbf{0.7553} & \textbf{.3649} & \textbf{.4985} & \textbf{.4213} & .2392 & --\\
\bottomrule
\end{tabular}
}
\caption{Consolidated results comparing fine-tuning XLM-R and zero-shot learning with GPT-4o. The table shows performance and cost comparisons between single-step and multi-step LLM prompting approaches, where the highest scores between these two approaches across all languages are \underline{underlined}. The top results across all three methods and languages are highlighted in \textbf{bold}.}
\label{tab:promptingmethods}
\end{table*}

Table \ref{tab:promptingmethods} consolidates the results from fine-tuning XLM-R and hierarchical zero-shot learning, showing that the multi-step approach achieves slightly better performance for main role prediction compared to the single-step approach. 
This can be attributed to the structured decomposition of the task. 

By explicitly isolating the main role prediction into a separate dedicated step, the model can focus on high-level role identification without the distraction of fine-grained details.
While this multi-step approach enhances performance for main role prediction, it underperforms compared to the single-step approach for predicting fine-grained roles. We hypothesize two potential reasons for this discrepancy:
\begin{enumerate}
\item \textbf{Error Propagation:} In the multi-step approach, the model first predicts the main role and then proceeds to predict fine-grained roles. Errors introduced during the main role prediction step can propagate to subsequent steps, thereby reducing the overall accuracy of fine-grained predictions.
\item \textbf{Loss of Joint Context:} The single-step approach enables the model to reason jointly about both main roles and fine-grained roles, allowing it to better capture dependencies between role labels. This integrated reasoning leads to more precise and consistent fine-grained predictions.
Another finding is that the multi-step approach is significantly more cost-effective than the single-step approach. This efficiency may stem from token efficiency, as multi-step prompts are designed to be more concise, with each step addressing a specific sub-task (e.g., main role followed by fine-grained role). Consequently, this results in fewer tokens per prompt.
\end{enumerate}

Table~\ref{tab:promptingmethods} additionally compares the performance of zero-shot approaches and XLM-R, both using similarly sized input contexts. We can see that XLM-R outperforms zero-shot methods on all evaluation measures except for Macro F1, where it shows the lowest performance among all approaches we experimented with. 

We hypothesize that this discrepancy arises because XLM-R had only seen a limited number of training instances for the rare roles, which makes it very hard for the model to learn to detect these categories. In contrast, zero-shot approaches do not rely on training data and thus are not constrained by this limitation. 

This hypothesis is based on observable patterns in the dataset and model performance. \mbox{Figure \ref{fig:fine_roles_histogram}} illustrates the fine-grained role distribution, revealing a significant imbalance, where we can see that certain roles are severely underrepresented. This imbalance intuitively contributes to the observed discrepancy between Micro and Macro F1 scores, as XLM-R's fine-tuning relies on sufficient training instances for rare roles, which are limited in this dataset. This observation is consistent with previous findings in NLP \cite{henning-etal-2023-survey}, where class imbalance is known to affect the performance for low-frequency classes.

In contrast, zero-shot approaches do not depend on labeled training data, which makes them less sensitive to class imbalance, which can explain their comparatively higher Macro F1 scores. While factors such as model architecture or prompt design could also influence performance, dataset imbalance remains a key explanatory factor supported by our analysis. Thus, while XLM-R achieves a higher Micro F1 score, the higher Macro F1 score that we observe for zero-shot approaches offers an intuitive explanation tied to the distribution of roles in the dataset.

\section{Conclusion and Future Work}
We presented a novel multilingual and hierarchical dataset for characterizing entity framing and role portrayal in news articles. The dataset introduces a unique taxonomy inspired by storytelling elements, featuring 22 fine-grained archetypes nested within three main categories: \emph{protagonist}, \emph{antagonist}, and \emph{innocent}. The dataset covers 1,378 recent news articles in five languages (Bulgarian, English, Hindi, European Portuguese, and Russian), spanning two domains: the Ukraine--Russia War and Climate Change. Over 5,800 entity mentions have been thoroughly annotated with role labels, capturing nuanced portrayals. We evaluated the dataset using fine-tuned state-of-the-art multilingual transformers and explored hierarchical zero-shot learning at the document, paragraph, and sentence level. Our experiments highlighted the potential of multilingual representations and hierarchical approaches for entity-framing tasks. We release the dataset to the community freely for research purposes, and we hope it will enable future research.

In future work, we plan to extend the annotations to additional languages and explore other sources, such as social media posts. This expansion aims to provide a broader understanding of role portrayal across diverse linguistic and contextual settings. We also intend to perform a more in-depth dataset analysis to examine formulations such as central entity identification. We further plan to evaluate the potential biases in the dataset aiming to ensure robustness for downstream tasks and end-user applications.

\section*{Limitations}

\paragraph{Corpus} Our dataset focuses on two domains: the Ukraine--Russia War and Climate Change, and covers news articles in five languages (Bulgarian, English, Hindi, Portuguese, and Russian). While these domains and languages provide a diverse foundation for entity framing analysis, the corpus contains 1,378 articles and it should not be considered representative of all news coverage or media landscapes in any specific country. Moreover, the dataset is not balanced with respect to topics, entities, or languages. Furthermore, human annotation of entity framing and role portrayal inevitably is subjective and annotators may subconsciously have biases that could influence quality. Despite providing detailed annotation guidelines and conducting quality control measures such as double annotation and adjudication through the curation process, some level of subjectivity may remain in the dataset. 

\paragraph{Baseline Models}
Our reported experiments utilize state-of-the-art baselines covering a range of fine-tuned multilingual transformer models and hierarchical zero-shot learning with LLMs. However, we have not yet explored alternative architectures or advanced techniques such as few-shot, instruction-based evaluation, or multitask learning. Future work could investigate these approaches to improve model efficiency and performance. 

Additionally, our zero-shot learning experiments rely on OpenAI's GPT-4o, a closed-source model that is subject to changes over time and may be deprecated in the future. This dependency may impact the reproducibility and interpretability. To address these challenges, future research should prioritize improving open-source models to ensure greater accessibility, transparency, and reproducibility.

\section*{Ethics and Broader Impact}

\paragraph{Biases}
Our dataset aims to capture a balanced range of perspectives. While our goal is to incorporate diverse news sources and viewpoints, achieving perfect balance is not always possible. Consequently, inherent biases in the original media sources may be present in the annotations. To reduce unwanted annotation biases, the corpus is annotated with clear instructions to annotators to focus strictly on the framing of entities, setting aside their personal opinions. All annotations are performed by subject-matter experts, and we did not use crowd-sourcing.

\paragraph{Intended Use and Misuse Potential}
The primary goal of this corpus is to facilitate research on entity framing, role portrayal, and media analysis. These tools can help researchers, journalists, and the general public identify framing patterns and biases in news content. However, there is a risk that the corpus could be misused for malicious purposes, such as manipulating news narratives. We urge users to employ this resource responsibly and remain aware of potential ethical risks associated with its misuse.

\paragraph{Environmental Impact}
The use of LLMs requires substantial computational power, contributing to carbon emissions. Even though we used LLMs in a zero-shot in-context-learning setting rather than training models from scratch, the LLMs still rely on GPUs for inference, which has an environmental impact.

\paragraph{Fairness}

Most of our annotators and curators come from the institutions of the co-authors of this manuscript and were fairly paid as part of their job duties. Few annotators were experienced analysts with full-time consulting roles and rates set by their contracting institutions. A fraction 
of the annotators were students from the respective
academic organizations. For two languages, a professional annotation company was contracted on rates based on country of residence. At the same time, some of the remaining annotators were researchers working primarily as linguists and lexicographers at their institute of affiliation and were all compensated according to local standards and their employment contracts.

\section*{Acknowledgments}

This research is partially funded by the EU NextGenerationEU, through the National Recovery and Resilience Plan of the Republic of Bulgaria, project SUMMIT, No BG-RRP-2.004-0008.

Giovanni Da San Martino is funded by the European Union under the National Recovery and Resilience Plan (NRRP), Mission 4 Component 2 Investment 1.3 - Call for tender No. 341 of March 15, 2022 of Italian Ministry of University and Research – NextGenerationEU; Code PE00000014, Concession Decree No. 1556 of October 11, 2022 CUP D43C22003050001, Progetto SEcurity and RIghts in the CyberSpace (SERICS) - Spoke 2 Misinformation and Fakes - DEcision supporT systEm foR cybeR intelligENCE (Deterrence). 

\bibliography{anthology,custom,shared}
\clearpage
\appendix
\section{Detailed Taxonomy with Examples}
\label{sec:detailed_taxonomy}

Below, we definer the entries in our taxonomy, and we give an example for each category. Any references to ``URW'' and ``CC'' below denote the Ukraine-Russia War, and the Climate Change domains, respectively.

\subsection{Protagonist}

\textbf{Guardian}: Heroes or guardians who protect values or communities, ensuring safety and upholding justice. They often take on roles such as law enforcement officers, soldiers, or community leaders (e.g., climate change advocacy community leaders). 
\\\underline{Example}: Police officers protecting citizens during a crisis, firefighters saving lives during a disaster, community leaders standing against crime or leaders standing up for action to address climate change. 

\textbf{Martyr}: Martyrs or saviors who sacrifice their well-being, or even their lives, for a greater good or cause. These individuals are often celebrated for their selflessness and dedication. This is mostly in politics, not in CC.
\\\underline{Example}: Civil rights leaders like Martin Luther King Jr., who was assassinated while fighting for equality, or journalists who risk their lives to report on corruption and injustice. 

\textbf{Peacemaker}: Individuals who advocate for harmony, working tirelessly to resolve conflicts and bring about peace. They often engage in diplomacy, negotiations, and mediation. This is mostly in politics, not in CC.
\\\underline{Example}: Nelson Mandela's efforts to reconcile South Africa post-apartheid, or diplomats working to broker peace deals between conflicting nations. 

\textbf{Rebel}: Rebels, revolutionaries, or freedom fighters who challenge the status quo and fight for significant change or liberation from oppression. They are often seen as champions of justice and freedom. 
\\\underline{Example}: Leaders of independence movements like Mahatma Gandhi in India, or modern-day activists fighting for democratic reforms in authoritarian regimes. In CC domain, this includes characters such as Greta Thunberg, or persons who, for instance, chain themselves to trees to prevent deforestation.

\textbf{Underdog}: Entities who are considered unlikely to succeed due to their disadvantaged position but strive against greater forces and obstacles. Their stories often inspire others. 
\\\underline{Example}: Grassroots political candidates overcoming well-funded incumbents, or small nations standing up to larger, more powerful countries. In CC, this could included NEs portrayed as underfunded organizations that are framed as showing promise to make positive impact on CC.

\textbf{Virtuous}: Individuals portrayed as virtuous, righteous, or noble, who are seen as fair, just, and upholding high moral standards. They are often role models and figures of integrity.
\\\underline{Example}: Judges known for their fairness, or politicians with a reputation for honesty and ethical behavior. In CC, this includes leaders standing up for environmental ethical values to protect planet Earth, or activists pushing for environmental sustainability.

\subsection{Antagonist}

\textbf{Instigator}: Individuals or groups initiating conflict, often seen as the primary cause of tension and discord. They may provoke violence or unrest.
\\\underline{Example}: Politicians using inflammatory rhetoric to incite violence, or groups instigating protests to destabilize governments. In CC, this could also include Greta Thunberg or activists chaining themselves to trees. In the previous example, they were portrayed in positive light as rebels. However, they could just as well be framed in a negative light if they are being portrayed as troublemakers and instigators of problems, and in such a scenario, they would also take the sub-role of Sabateur.

\textbf{Conspirator}: Those involved in plots and secret plans, often working behind the scenes to undermine or deceive others. They engage in covert activities to achieve their goals.
\\\underline{Example}: Figures involved in political scandals or espionage, such as Watergate conspirators or modern cyber espionage cases. In CC, this could manifest as persons or organizations conspiring to bypass environmental regulations for profit. 

\textbf{Tyrant}: Tyrants and corrupt officials who abuse their power, ruling unjustly and oppressing those under their control. They are often characterized by their authoritarian rule and exploitation.
\\\underline{Example}: Dictators like Kim Jong-un in North Korea, or corrupt officials embezzling public funds and suppressing dissent. 

\textbf{Foreign Adversary}: Entities from other nations or regions creating geopolitical tension and acting against the interests of another country. They are often depicted as threats to national security. This is mostly in politics, not in CC.
\\\underline{Example}: Rival nations involved in espionage or military confrontations, such as the Cold War adversaries, or countries accused of election interference. In CC, foreign adversaries could include portrayal of how other countries are not adhering to CC policies (e.g., China refuses to adhere to CC policies resulting in 20\% increase in CO2 emissions.

\textbf{Traitor}: Individuals who betray a cause or country, often seen as disloyal and treacherous. Their actions are viewed as a significant breach of trust. This is mostly in politics, not in CC.
\\\underline{Example}: Whistleblowers revealing sensitive information for personal gain, or soldiers defecting to enemy forces. Note that if whistleblowers are portrayed in a positive light, their role would be Virtuous. This could equally apply to both politics and CC.

\textbf{Spy}: Spies or double agents accused of espionage, gathering and transmitting sensitive information to a rival or enemy. They operate in secrecy and deception. This is mostly in politics, not in CC.
\\\underline{Example}: Historical figures like Aldrich Ames, who spied for the Soviet Union, or contemporary cases of corporate espionage.

\textbf{Saboteur}: Saboteurs who deliberately damage or obstruct systems, processes, or organizations to cause disruption or failure. They aim to weaken or destroy targets from within.
\\\underline{Example}: Insiders tampering with critical infrastructure, or activists sabotaging industrial operations.

\textbf{Corrupt}: Individuals or entities that engage in unethical or illegal activities for personal gain, prioritizing profit or power over ethics. This includes corrupt politicians, business leaders, and officials.
\\\underline{Example}: Companies involved in environmental pollution, executives engaged in massive financial fraud, or politicians accepting bribes and engaging in graft.

\textbf{Incompetent}: Entities causing harm through ignorance, lack of skill, or incompetence. This includes people committing foolish acts or making poor decisions due to lack of understanding or expertise. Their actions, often unintentional, result in significant negative consequences.
\\\underline{Example}: Leaders making reckless policy decisions without proper understanding, officials mishandling crisis responses, or managers whose poor judgment leads to organizational failures.
\newpage

\textbf{Terrorist}: Terrorists, mercenaries, insurgents, fanatics, or extremists engaging in violence and terror to further ideological ends, often targeting civilians. They are viewed as significant threats to peace and security. This is mostly in politics, not in CC.
\\\underline{Example}: Groups like ISIS or Al-Qaeda carrying out attacks, or lone-wolf terrorists committing acts of violence.

\textbf{Deceiver}: Deceivers, manipulators, or propagandists who twist the truth, spread misinformation, and manipulate public perception for their own benefit. They undermine trust and truth.
\\\underline{Example}: Politicians spreading false information for political gain, or media engaging in propaganda.

\textbf{Bigot}: Individuals accused of hostility or discrimination against specific groups. This includes entities committing acts falling under racism, sexism, homophobia, antisemitism, islamophobia, or any kind of hate speech. This is mostly in politics, not in CC.

\subsection{Innocent}

\textbf{Forgotten}: Marginalized or overlooked groups who are often ignored by society and do not receive the attention or support they need. This includes refugees, who face systemic neglect and exclusion.
\\\underline{Example}: Indigenous populations facing ongoing discrimination; homeless individuals struggling without adequate support; refugees fleeing conflict or persecution.

\textbf{Exploited}: Individuals or groups used for others' gain, often without their consent and with significant detriment to their well-being. They are often victims of labor exploitation, trafficking, or economic manipulation.
\\\underline{Example}: Workers in sweatshops; victims of human trafficking; communities suffering from corporate exploitation of natural resources.

\textbf{Victim}: People cast as victims due to circumstances beyond their control, specifically in two categories: (1) victims of physical harm, including natural disasters, acts of war, terrorism, mugging, physical assault, ... etc., and (2) victims of economic harm, such as sanctions, blockades, and boycotts. Their experiences evoke sympathy and calls for justice, focusing on either physical or economic suffering.
\\\underline{Example}: Victims of natural disasters, such as hurricanes or earthquakes; individuals affected by violent crimes. Victims of economic blockades, sanctions, or boycotts.

\textbf{Scapegoat}: Entities blamed unjustly for problems or failures, often to divert attention from the real causes or culprits. They are made to bear the brunt of criticism and punishment without just cause.
\\\underline{Example}: Minority groups blamed for economic problems; political opponents, accused of provoking national strife, without evidence.

\section{Annotation Guidelines}
\label{sec:annotation_guidelines}

These guidelines aim to prepare the annotators and avoid human biases before starting the annotation:

\begin{itemize}
    \item The annotators should get acquainted with the two domains covered by the tasks; for instance, ~\cite{cc-denial-tax} and ~\cite{illiberalism} provide a good coverage of the CC and URW domains, respectively,
    \item The annotators' opinions on the topics and sympathies towards key entities mentioned in the articles are irrelevant and should by no means impact the annotation process and their choices, 
    \item The annotators should not exploit any external knowledge bases for the purpose of annotating documents.
\end{itemize}

Our guidelines for annotating and curating the entity framing corpus are as follows.
Any references to ``URW'' and ``CC'' below denote the Ukraine-Russia War, and the Climate Change domains, respectively.

\begin{enumerate}

    \item The entities of interest are understood in a broad sense to include both traditional named entities (such as persons, organizations, and locations) and {\em toponym-derived} entities. Toponym-derived entities are phrases that indicate a group or collective identity based on a place or affiliation, including, but not limited to:
    \begin{itemize}
        \item Political, military, or social groups defined by their association with a location or entity, e.g., ``Trump supporters,'' or ``residents of Ukraine.''
        \item Entities denoting a geographic or organizational affiliation, such as ``Russian forces'' or ``European officials.''
    \end{itemize}

    \item Annotators are provided with a number of news articles and are expected to assign roles to named entities that are \textbf{central} to the article's story, according to the taxonomy of roles that was provided earlier. 
    \item Annotators are provided with a detailed taxonomy that includes definitions and examples. 
    \item The title of an article should not be annotated. The title of the article is the first block of text that appears in the annotation platform {\em INCEpTION}.
    \item Only named entities that are central to the narrative of the article should be annotated. Unnamed entities (i.e., nominal entity mentions such as ``migrants'') should not be annotated. 

For more details on what qualifies as a named entity, in addition to the definition of the broader sense of named entities given above in these guidelines, the annotators should also examine the NER annotation guidelines in \href{http://www.universalner.org/guidelines/}{www.universalner.org/guidelines/}.
    
    \item Annotators pick one or more fine-grained roles for the named entities they believe are central to the article's story. 
    \item Entity mentions can be assigned fine-grained roles from more than one main role. However, during curation, we will not include these instances in the current version of the corpus, even though we annotate them.
    \item Named entities that are not central to the story should not be annotated.

The determination of how central a named entity is in an article is admittedly subjective. To reduce bias, such determination should be based on the careful reading of the article. 
    
    \item As a general rule, annotators should annotate only the first mention of each entity where it is clear that this entity has the specific role(s). There is no need to annotate subsequent mentions of this entity with the same role, but annotating more mentions with the same surface form and role is not a mistake; it is simply not required.

    This rule also extends to surface mentions of the same entity. For example, ``Putin'' and ``Vladimir Putin'' are both surface mentions of the same entity, so only the first occurrence would be annotated.

    On the other hand, while entities such as ``Moscow'', ``Russia'', and ``Putin'' are closely related, they are not surface forms of the same entity, and are considered to be distinct separate entities.


    \item If the above results in more than one mention of the same entity with the same role, the curator does not need to remove all of these additional mentions. We keep all of them.
    
    \item Should an entity that was previously annotated with a certain role appear in a different context with different roles, the first mention where the roles changed should be annotated.

The above rule is repeated as many times as the entity changes roles across mentions.  For example, if an entity, let's say NATO, appears 20 times in an article, the first 10 mentions show NATO as a Guardian and a Virtuous entity.  The 11th-15th mentions portray NATO as a Foreign Adversary, and the 16th-20th mentions portray NATO as Exploited, then we need only 3 annotations in total to account for the 3 different roles that NATO was portrayed as.  These 3 annotations should all be the first mentions where NATO assumed each distinct role (i.e., mention 1, mention 11, and mention 16 should be annotated).
    \item If different surface forms for the same named entity (e.g., NATO vs.~North Atlantic Treaty Organization) appear in the article, it is sufficient to annotate only one of the surface forms.
    \item If the above results in multiple surface forms of the same entity being annotated, the curator does not need to remove all of these additional mentions. We keep all of them.
    \item There is no ``Other'' label in the taxonomy, as mentions without a discernible role in relation to the taxonomy are simply not assigned any role.
    \item The curator may see conflicting annotations in the curation mode and can resolve the conflict, and then the remaining non-conflicting roles can be checked and adapted accordingly. 

\end{enumerate}

\newpage
\section{Experimental Settings}
\label{sec:appendix_experiments}

All fine-tuning experiments were conducted on a single NVIDIA RTX 4090 GPU with 24 GB of memory. We fine-tuned XLM-R (XLM-RoBERTa) in a single run, using a fixed random seed to ensure reproducibility. When the input context was at the sentence granularity, we performed sentence splitting using Stanza pipelines for each one of our five languages. For XLM-R, default settings were applied, with the following configurations:

\begin{itemize}
    \item Model: XLM-R\textsubscript{base} (125M parameters) 
    \item Learning Rate: 2e-5
    \item Batch Size: 8
    \item Epochs: 20 (with early stopping of 3 based on validation loss)
    \item Random Seed: 42
    \item Weight Decay: 0.01
\end{itemize}

To optimize performance, the sigmoid thresholds for fine-grained role predictions were tuned on the validation set. These optimized thresholds were then applied to generate predictions on the test set.




For the zero-shot experiments, we used OpenAI's GPT-4o (gpt-4o-2024-11-20) with a temperature setting of 0.2 to produce more conservative responses. To ensure the outputs conformed to our defined data types, we employed OpenAI's Structured Outputs API, which returned results in the expected JSON format.

\clearpage
\onecolumn

\section{Dataset Statistics}

\label{sec:appendix_stats}

\begin{figure*}[!h]
    \centering
    \includegraphics[width=1\textwidth]{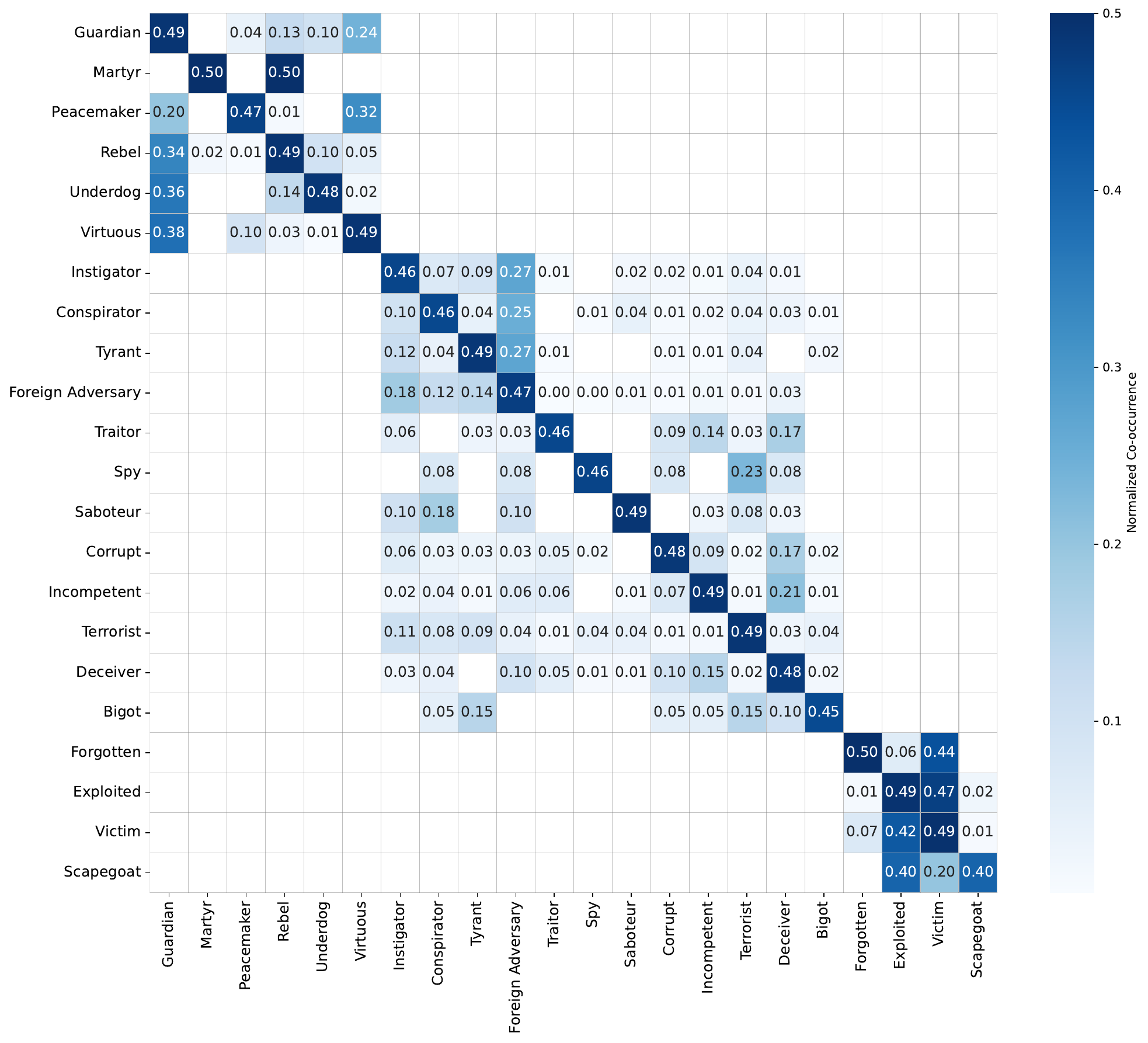}
    \caption{Normalized co-occurrence of fine-grained roles.}
    \label{fig:fine_roles_co_occurence_normalized}
\end{figure*}

\begin{figure}[!h]  
    \centering

    \begin{subfigure}[t]{0.9\textwidth}  
        \centering
        \includegraphics[width=\textwidth]{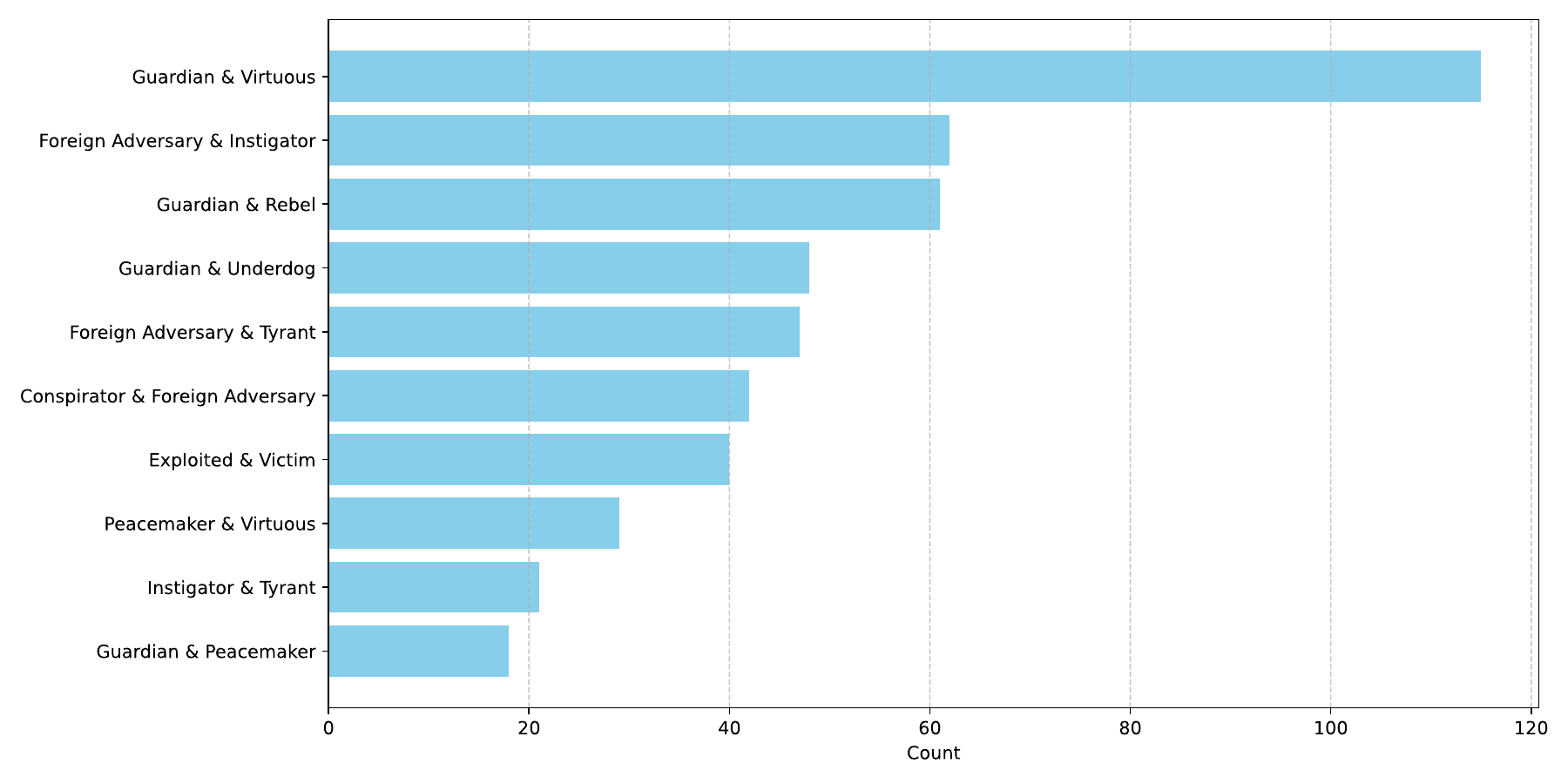}
        \caption{}
        \label{fig:frequent_pairs}
    \end{subfigure}
    \hfill  
    
    \begin{subfigure}[t]{\textwidth}  
        \centering
        \includegraphics[width=\textwidth]{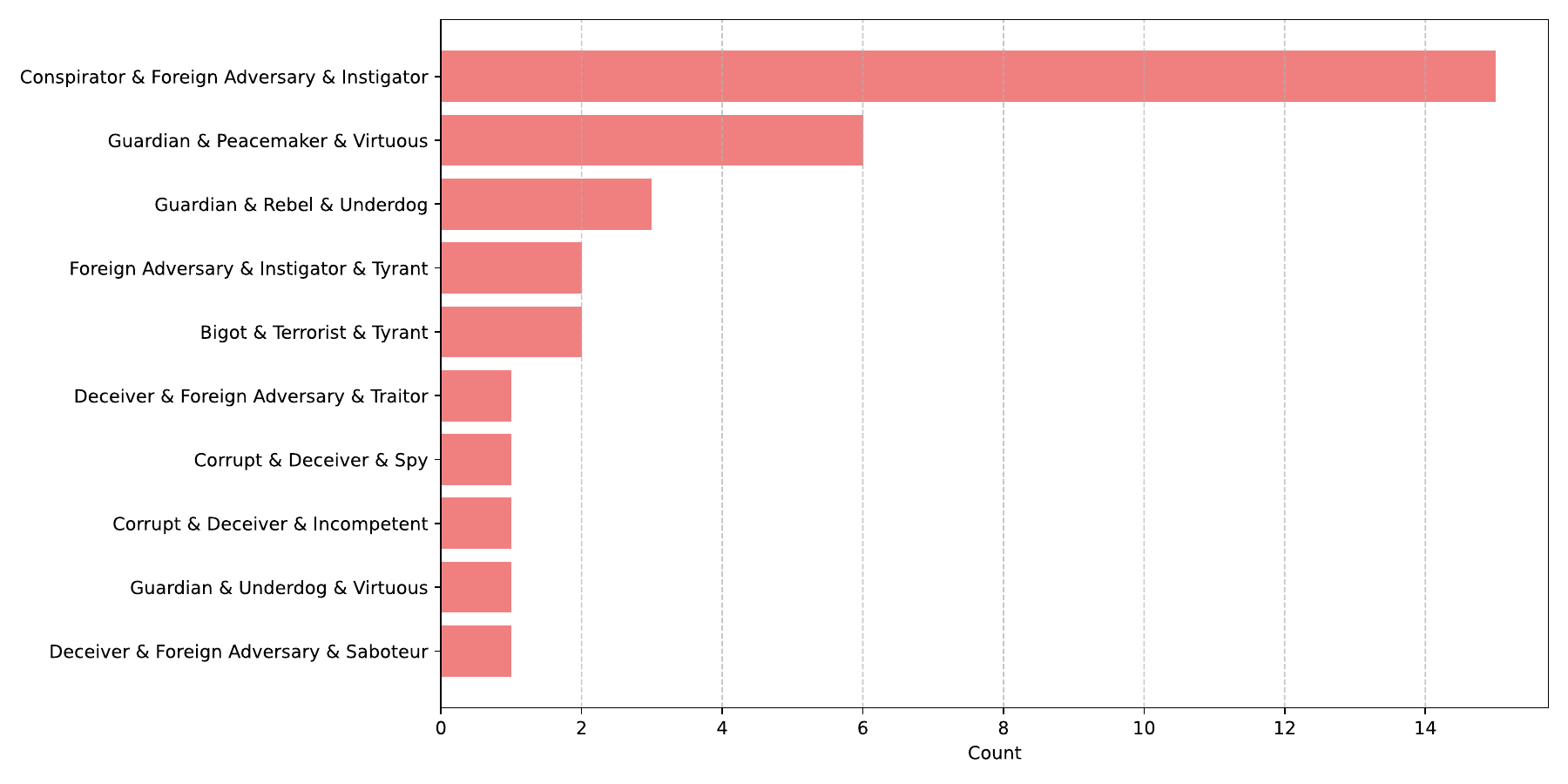}
        \caption{}
        \label{fig:frequent_triplets}
    \end{subfigure}

    \caption{The 10 most frequent co-occurring (a) pairs and (b) triplets of fine-grained roles.}
    \label{fig:frequent_roles_combined}
\end{figure}

\begin{figure}[!h]  
    \centering

    \begin{subfigure}[t]{0.9\textwidth}  
        \centering
        \includegraphics[width=\textwidth]{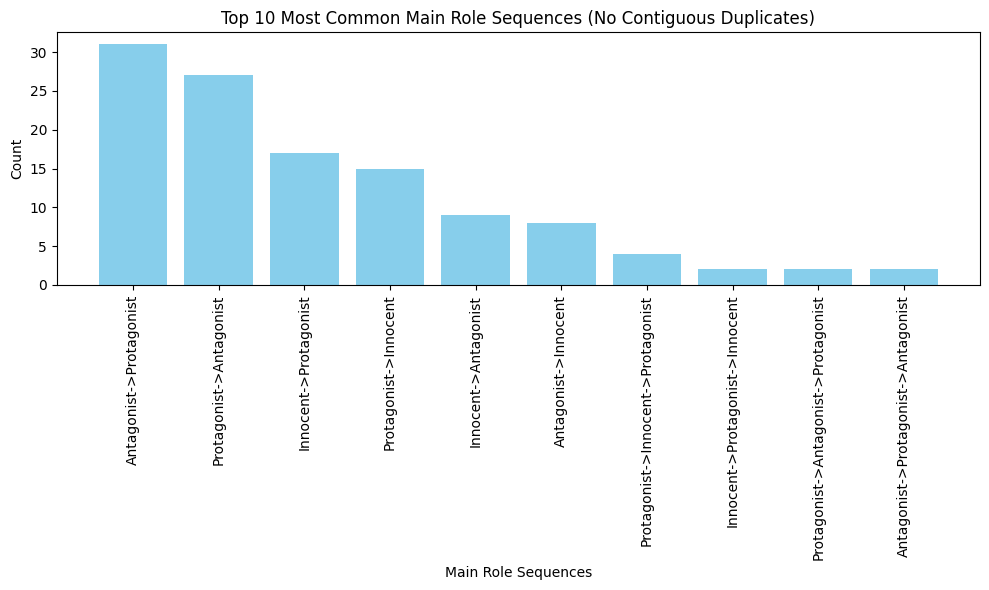}
        \caption{}
        \label{fig:main_role_seq}
    \end{subfigure}
    \hfill  
    
    \begin{subfigure}[t]{\textwidth}  
        \centering
        \includegraphics[width=0.9\textwidth]{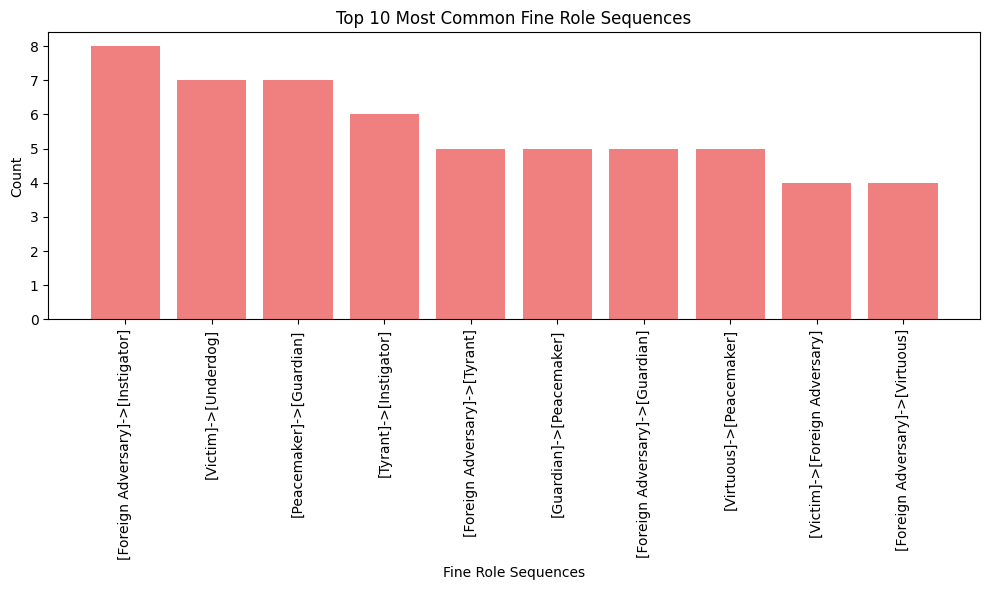}
        \caption{}
        \label{fig:fine_role_seq}
    \end{subfigure}

    \caption{The 10 most frequent (a) main role and (b) fine-grained role transition sequences.}
    \label{fig:transition_sequences}
\end{figure}

\begin{figure}  

    \begin{subfigure}[b]{0.9\textwidth}  
        \centering
        \includegraphics[width=\textwidth]{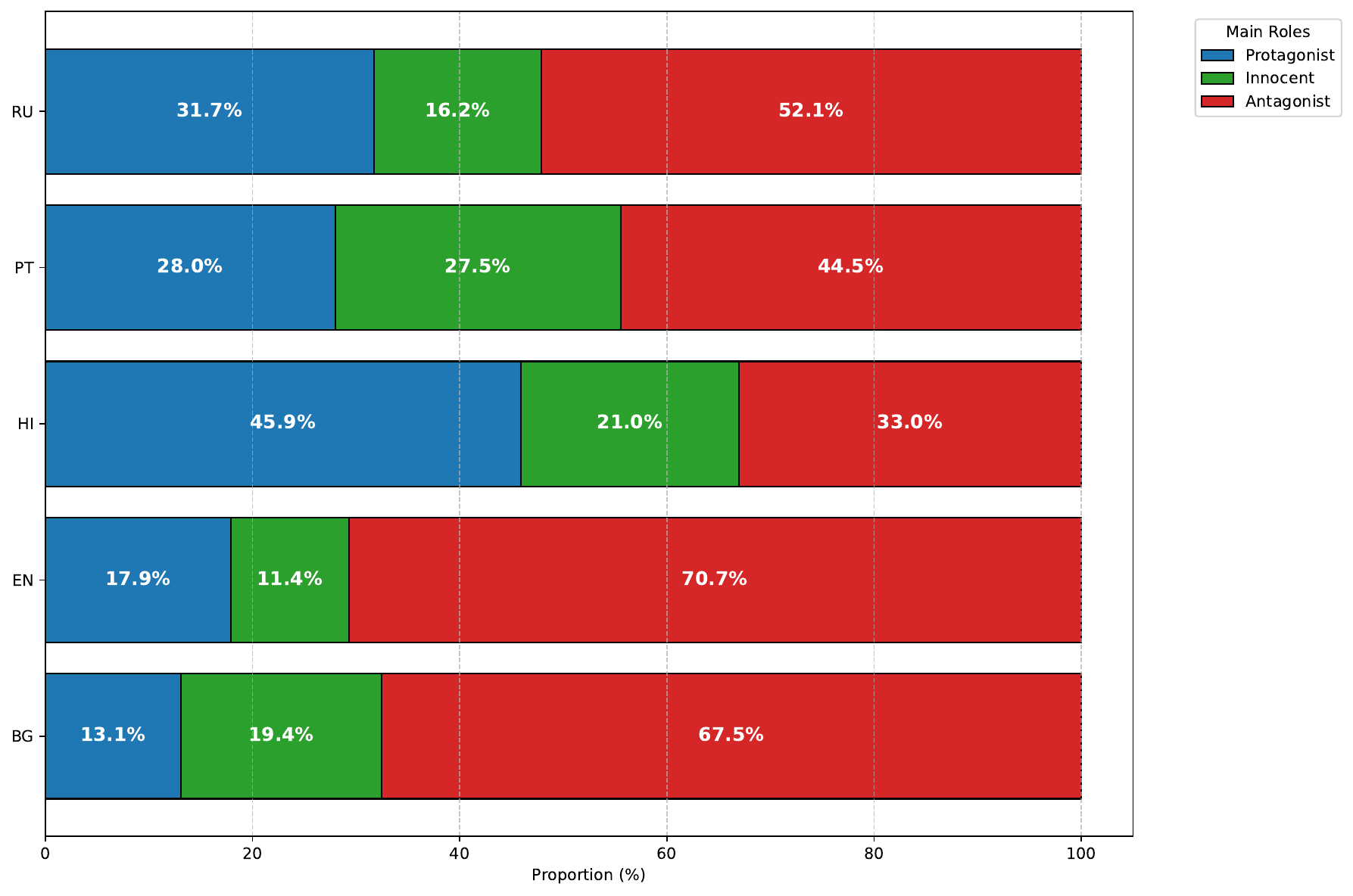}
        \caption{}
        \label{fig:main_roles_per_language}
    \end{subfigure}
    
    \begin{subfigure}[b]{\textwidth}  
        \centering
        \includegraphics[width=\textwidth]{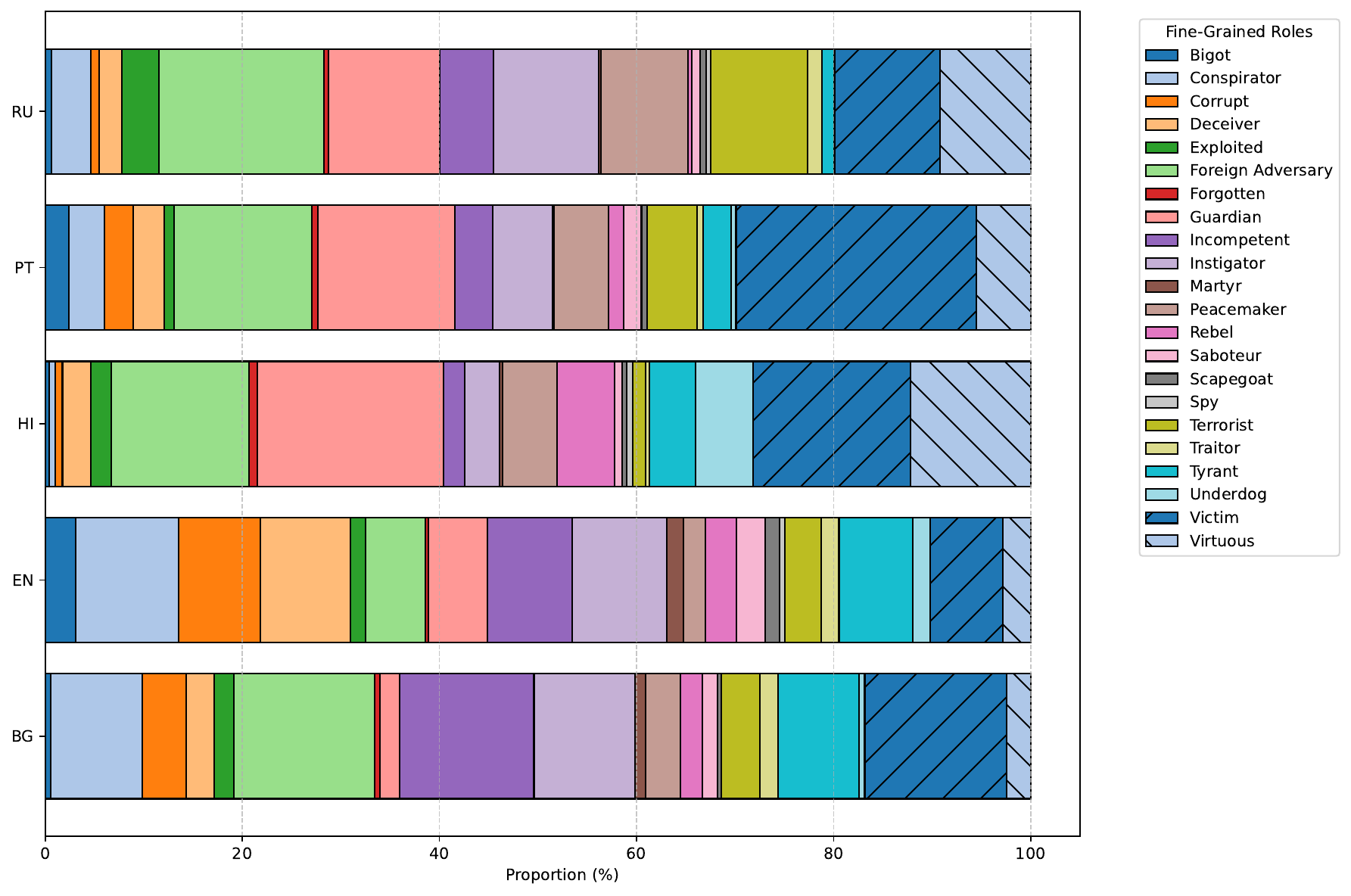}
        \caption{}
        \label{fig:fine_roles_per_language}
    \end{subfigure}

    \caption{Proportions of (a) main roles and (b) fine-grained roles per language.}
    \label{fig:proportions_of_roles}
\end{figure}

\begin{figure}
    \centering
    \includegraphics[scale=1.5]{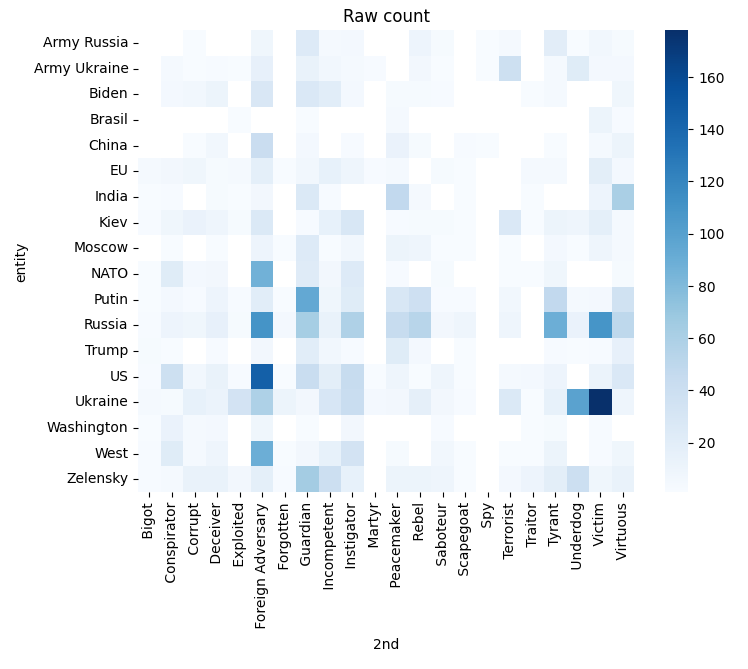}
    \caption{Heathmap of the raw count of mention of entity and 2nd level role, where entities are defined and selected as in Figure~\ref{fig:hist_ent}}
    \label{fig:entity_2nd}
\end{figure}

\begin{figure}
    \centering
    \includegraphics[scale=0.22]{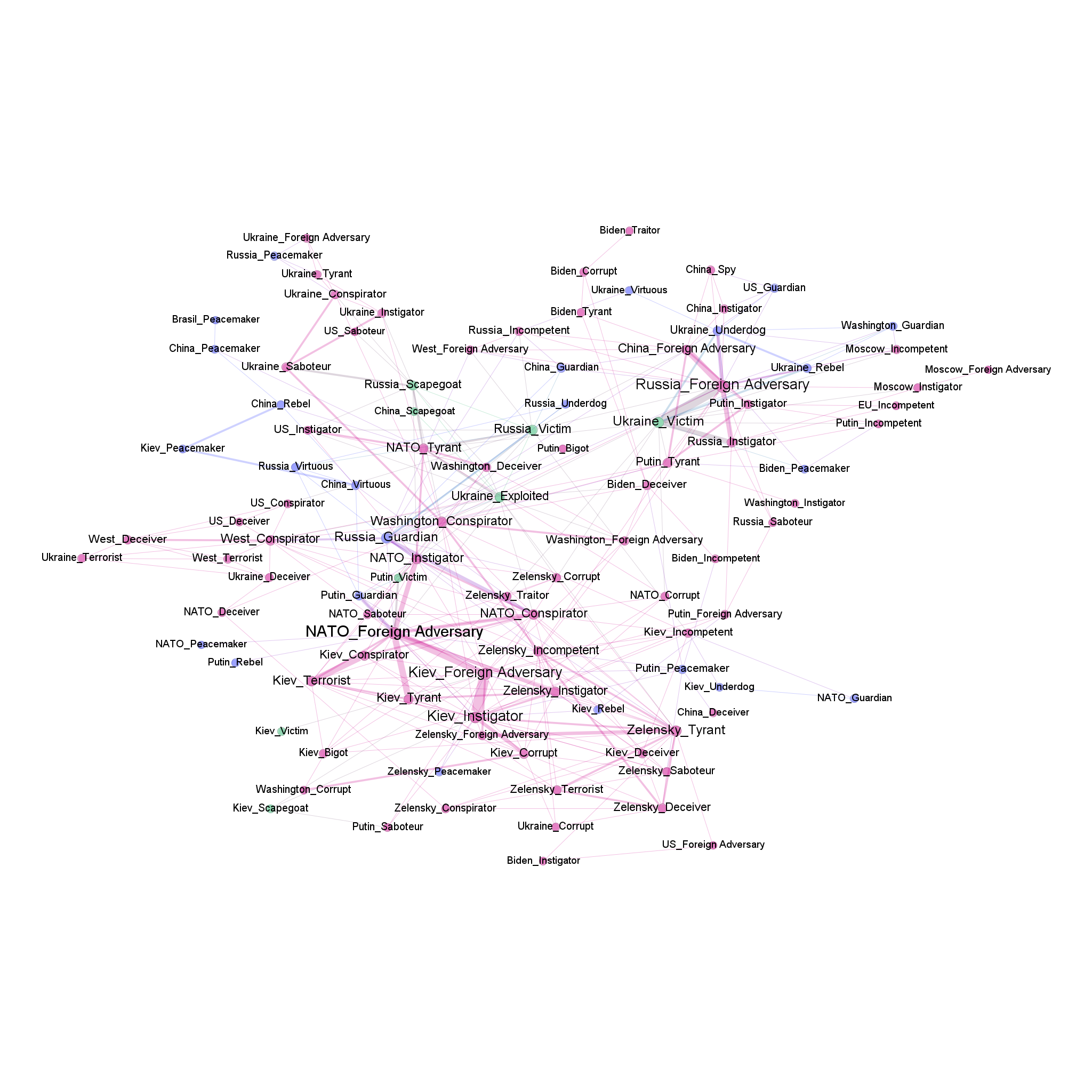}
    \caption{Graph of the top entities, nodes are a pair of entity and 2nd level role, node size is relative to the count of mentions, node color codes the 1st level role, there is and edge between two nodes, if they appear in the same document. This graph illustrate how group of entity+role pairs can be used to identify potential narratives.}
    \label{fig:enter-label}
\end{figure}

\clearpage
\section{Prompts for Hierarchical Zero-Shot Experiments}
\label{sec:appendix_zeroshot}

\begin{figure*}[htbp]
\centering
\begin{framed}
\begin{minipage}{0.9\textwidth}
\begin{Verbatim}[fontsize=\scriptsize,commandchars=\\\{\}]

\lightbrowntext{You are an expert at identifying entity framing and role portrayal in news articles. Analyze the following entity} 
\lightbrowntext{mention in context, and predict its main role and fine-grained role(s) from the taxonomy below.} 

\lightbrowntext{Taxonomy:} \darkbluetext{\{}\lightbluetext{detailed taxonomy with definitions and examples}\darkbluetext{\}}

\lightbrowntext{Context Around Entity:} \darkbluetext{\{}\lightbluetext{context}\darkbluetext{\}}

\lightbrowntext{Entity Mention:} \darkbluetext{\{}\lightbluetext{entity mention}\darkbluetext{\}}

\lightbrowntext{Task: Based on the provided context, assign to the entity mention at least one fine-grained role and}
\lightbrowntext{exactly one main role.}

\lightbrowntext{Return a JSON that has below attributes:}
\lightbrowntext{- \textbf{main role}: either one of Protagonist, Antagonist, or Innocent}
\lightbrowntext{- \textbf{fine grained roles}: a list of all your predicted fine-grained roles}

\end{Verbatim}
\end{minipage}
\end{framed}
\caption{\textbf{Single-Step Prompt Template.} The detailed taxonomy is the same one shown in \ref{sec:detailed_taxonomy}. The context is the text consisting of entity mention along with the 20 words before and after the entity mention.}
\label{fig:single_step_prompt_template}
\end{figure*}

\begin{figure*}[htbp]
\centering
\begin{framed}
\begin{minipage}{0.9\textwidth}
\begin{Verbatim}[fontsize=\scriptsize,commandchars=\\\{\}]
\darkbluetext{First Step (LLM Call 1): Predict the Main Role}

\lightbrowntext{You are an expert at identifying entity framing and role portrayal in news articles. Analyze the following entity }
\lightbrowntext{mention in context, and predict its main role from the taxonomy below.} 

\lightbrowntext{Taxonomy:} \darkbluetext{\{}\lightbluetext{list of fine-grained roles per main role}\darkbluetext{\}}

\lightbrowntext{Context Around Entity:} \darkbluetext{\{}\lightbluetext{context}\darkbluetext{\}}

\lightbrowntext{Entity Mention:} \darkbluetext{\{}\lightbluetext{entity mention}\darkbluetext{\}}

\lightbrowntext{Task: Based on the provided context, assign to the entity mention exactly one main role.}

\lightbrowntext{Return a JSON that has this attribute:}
\lightbrowntext{- \textbf{main role}: either one of Protagonist, Antagonist, or Innocent}


\darkbluetext{Second Step (LLM Call 2): Predict the Fine-Grained Role}

\lightbrowntext{You are an expert at identifying entity framing and role portrayal in news articles. This entity is}
\lightbrowntext{portrayed as a(n)} \darkbluetext{\{}\lightbluetext{main role}\darkbluetext{\}}\lightbrowntext{ and your task is to analyze the entity mention in context}
\lightbrowntext{and predict its fine-grained role(s) from the taxonomy below.}

\lightbrowntext{Taxonomy:} \darkbluetext{\{}\lightbluetext{pertinent portion of the detailed taxonomy with definitions and examples}\darkbluetext{\}}

\lightbrowntext{Context Around Entity:} \darkbluetext{\{}\lightbluetext{context}\darkbluetext{\}}

\lightbrowntext{Entity Mention:} \darkbluetext{\{}\lightbluetext{entity mention}\darkbluetext{\}}

\lightbrowntext{Task: Based on the provided context, assign to the entity mention at least one fine-grained role.}

\lightbrowntext{Return a JSON that has this attribute:}
\lightbrowntext{- \textbf{fine grained roles}: a list of all your predicted fine-grained roles}
\end{Verbatim}
\end{minipage}
\end{framed}
\caption{\textbf{Multi-Step Prompt Template.} In the first step, the taxonomy is only the tree structure of the taxonomy and does not include any definitions or examples. In the second step, the detailed taxonomy only includes the branch under the predicted main role in the first step. The context is as defined in Figure \ref{fig:single_step_prompt_template}. }
\label{fig:multi_step_prompt_template}
\end{figure*}

\clearpage
\section{Annotation Tool}
\label{sec:annotation_tool}
We used the Inception~\cite{tubiblio106270} platform\footnote{https://inception-project.github.io/} to annotate our corpus because it has a rich set of features that extends beyond mere annotation to also include useful tools such as the ability to perform annotation adjudication through curation, monitoring the annotation progress, and calculating agreement between annotators. Inception allows to assign the following roles to users: annotator, curator, and manager.

\begin{figure}[!htpb]
\centering
\includegraphics[width=1\textwidth]{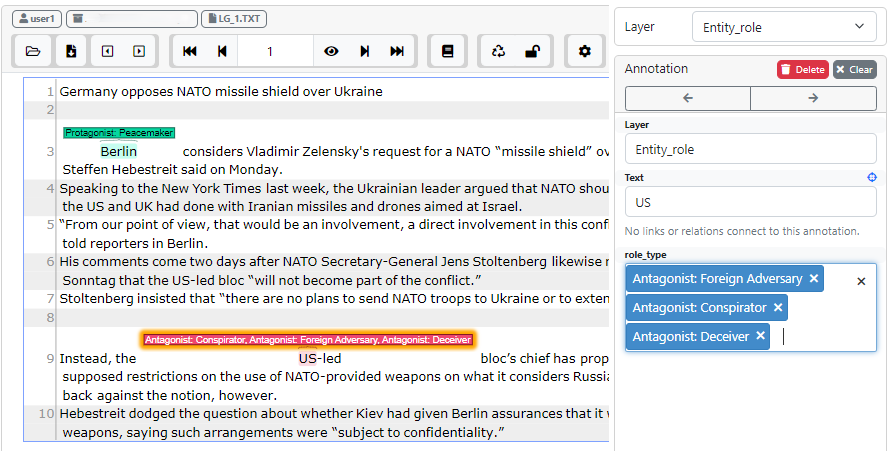}
\caption{Annotating entity framing using Inception.}
\label{fig:INC_entity}
\end{figure}

To annotate a mention of an entity with a role, annotators should go to the part of the article where the entity is mentioned and select it. After selecting an entity mention, annotators can then assign roles as shown in \ref{fig:INC_entity}.

\end{document}